# Structural damage detection via hierarchical damage information with volumetric assessment

Isaac Osei Agyemang[a,b]*, Isaac Adjei-Mensah[b], Daniel Acheampong[c], Gordon Owusu Boateng[b], Adu Asare Baffour[d]

[a] *Data Science and Big Data Technology, Stirling College, Chengdu University, Chengdu 610054, China*
[b] *School of Information and Communication Engineering, University of Electronic Science and Technology of China, Chengdu 610054, China*
[c] *Lutgert College, Florida Gulf Coast University, USA*
[d] *School of Science and Engineering, University of Missouri-Kansas City, USA*

**Abstract:** Structural health monitoring (SHM) is essential for ensuring the safety and longevity of infrastructure, but complex image environments, noisy labels, and reliance on manual damage assessments often hinder its effectiveness. This study introduces the Guided Detection Network (Guided-DetNet), a framework designed to address these challenges. Guided-DetNet is characterized by a Generative Attention Module (GAM), Hierarchical Elimination Algorithm (HEA), and Volumetric Contour Visual Assessment (VCVA). GAM leverages cross-horizontal and cross-vertical patch merging and cross-foreground-background feature fusion to generate varied features to mitigate complex image environments. HEA addresses noisy labeling using hierarchical relationships among classes to refine instances given an image by eliminating unlikely class instances. VCVA assesses the severity of detected damages via volumetric representation and quantification leveraging the Dirac delta distribution. A comprehensive quantitative study and two robustness tests were conducted using the Pacific Earthquake Engineering Research (PEER) Hub Image-Network dataset, and a drone-based application, which involved a field experiment, was conducted to substantiate Guided-DetNet's promising performances. In triple classification tasks, the framework achieved 96% accuracy, surpassing state-of-the-art classifiers by up to 3%. In dual detection tasks, it outperformed competitive detectors with a precision of 94% and a mean average precision (mAP) of 79% while maintaining a frame rate of 57.04 frames-per-second, suitable for real-time applications. Additionally, robustness tests demonstrated resilience under adverse conditions, with precision scores ranging from 79% to 91%. Guided-DetNet is established as a robust and efficient framework for SHM, offering advancements in automation and precision, with the potential for widespread application in drone-based infrastructure inspections.

## 1. Introduction

Structural Health Monitoring (SHM) plays a pivotal role in ensuring the safety and longevity of civil infrastructures (e.g., buildings, dams, tunnels, bridges, roads) [1]-[2]. Structural health monitoring has experienced significant progress due to the integration of deep learning methods [3]–[7], specifically Convolutional Neural Networks (CNNs), in object detection. Deep Convolutional Neural Networks (DCNNs), specifically object detectors, have exhibited satisfactory capabilities in identifying and localizing structural damages, thereby transforming computer vision-based techniques for detecting structural damages. Nevertheless, the effectiveness of object detectors is significantly impacted by the quality of the data on which they are trained, especially when identifying structural damages compared to other related object detection tasks [8]-[9].

Structural damage detection presents distinct challenges compared to general object detection tasks in typical environments. Unlike general object detection scenarios, where objects are often captured in controlled settings or well-defined backgrounds, structural damage detection must operate in complex and dynamic environments. These include occlusions caused by structural elements, vegetation, or other obstructions, background variations due to diverse surface textures (e.g., concrete, metal, or brick), and variability in lighting and weather conditions (e.g., fog, snow, or rain) [10]. These constraints are evident within the domain of structural health monitoring, where there is a notable variability in environmental conditions. This variability challenges the robustness of object detectors when deployed in various structural health monitoring environments and beyond. Additionally, one of the primary challenges to structural damage detection is noisy labels of images, which greatly hinders the detection of structural damage. The intrinsic complexity and diversity of structural damages, such as size, shape, and severity, lead to inaccurate annotation/labeling (noisy labeling), hindering object detectors' training [11]; this is less common in general object detection tasks, which typically deal with well-defined and standardized categories (e.g., cars, people, animals) [12]–[14]. The presence of noisy labeling not only undermines the dependability of the training process but also impedes the accuracy of subsequent detection, presenting a significant challenge to the practical implementation of object detectors in real-time applications in SHM. Also, the efficacy of object detectors is compromised in complex image environments [15]-[16].

---

*Corresponding author: Stirling College, Chengdu University.
Email address: isaac@stir.cdu.edu.cn or i.o.agyemang@stir.ac.uk (I.O. Agyemang)

Further, unlike general object detection, structural damage detection best requires leveraging hierarchical relationships between damage types (e.g., crack type, severity level, and affected structural components); this hierarchical complexity is rarely addressed in generic object detection frameworks. Not using hierarchical information associated with structural damage detection undermines the performance of object detectors, potentially being one of the reasons for false detections. The absence of object detectors or frameworks that utilize well-defined hierarchical structural damage detection datasets limits the capacity of object detectors to identify structural damages effectively. It is of utmost importance to acknowledge and rectify this deficiency to improve the understanding of detection outcomes and offer practical knowledge for making informed decisions in structural health monitoring. Besides, using hierarchical information associated with structural damages can potentially mitigate noisy labeling. General object detection systems are primarily focused on identifying and localizing objects. However, automated SHM applications seek autonomy regarding damage detection and assessment post-detections [17]–[19]; however, the focus has mostly been on the detection aspect, leaving out the assessment. This has led to the requirement of civil practitioners for judgment and analysis post-damage detection, which is also a continuous challenge in automated structural health monitoring. It is essential to have frameworks with features of autonomous assessments post-structural damage detection. These highlight the urgent requirement for transformative frameworks in automated structural damage detection and assessment.

To this end, this study mitigates the challenge related to noisy labels in structural damage detection, complex image environments, and damage assessment post-detections via the proposition of a Guided-DetNet. The Guided-DetNet framework comprises knowledge of hierarchical structural damage information, a robust attention mechanism with multi-varied generative feature maps, and volumetric contour visual assessments. The main contributions of the study are as follows:

1. The proposed Guided-DetNet is a multi-tasking framework comprising a classifier and detector for triple classification and a dual detection task based on hierarchical information, which utilizes the proposed hierarchical elimination algorithm.
2. A generative attention module (GAM), which introduces multi-varied dynamic feature descriptors, is proposed. GAM induces multi-varied dynamic features, which include refined features, refined feature plus, cross-horizontal and cross-vertical patch merge features, and foreground-background fused features, all to strengthen the robustness of Guided-DetNet.
3. A damage volumetric contour visual assessment module is proposed to quantify and determine the surface condition of an inspected civil infrastructure post-detection.
4. The application scenario of Guided-DetNet is demonstrated in a visual-based drone structural damage detection task.

The paper is structured with the literature review and proposed Guided-DetNet in Sections 2 and 3, respectively. Sections 4 and 5 detail the study's experiment setup, as well as the results and discussions, respectively. The conclusion of the study is given in Section 6.

## 2. Literature review
### 2.1. Structural damage detection and assessment via deep learning

Structural health monitoring applications are designed to inspect, assess, and evaluate civil infrastructure health statues. In most cases, SHM applications operate based on one or a mix of the three common computer vision-based tasks: classification, detection, and segmentation.

Regarding some of the current classification-based SHM methods, an optimized mansory façade classifier was proposed based on transfer learning and limited data [20]. A specialized curated dataset was used to train the optimized mansory façade classifier, which makes the classifier tailored for specific tasks; hence, generalization capabilities are limited. Similarly, transfer learning was adapted in the proposition of a hybrid CNN classifier, which utilizes AlexNet [21] and a threshold technique in concrete crack classification [22]. The extracted features leverage the threshold technique prior to the classification of the cracks. StairNet [23], an extension of EfficientNetV2 [24], was designed to focus on the feature contribution of the early, middle, and late blocks in a concrete crack-based classification study. The study's findings showed that the early, middle, and late blocks of the StairNet attained varying accuracy based on variant structural features. A binary classifier for crack classification and depth measurement was proposed by leveraging CNN and regression models [25]. The CNN was used for classification; afterward, the extracted features were fed to two regression models (XGBoost and Random Forest) for crack depth measurement. Similarly, a proposed approach for classifying the severity of concrete spalls utilized Deep Convolutional Neural Networks (DCNN) and an Extreme Gradient Boosting Machine (XGBoost) [26]. Feature extraction approaches such as center symmetric local binary pattern and local binary pattern were employed as texture descriptors to delineate the characteristics of a concrete surface susceptible to spalling. Also, a probabilistic hybrid model for surface concrete damage classification was proposed by leveraging Bayesian inference embedded in a deep convolutional neural network [27]. The model parameters in the probability layer were altered from deterministic values to Gaussian distributions. Other notable SHM classification studies include a one-dimensional CNN, BuildingNet, focusing on model optimization for damage classification [28].

In contrast to the classification-based SHM, detection is another vision-based task in SHM. Most recently, a YOLOv5 detector, which utilizes semantic features of street-view images, was proposed for large-scale detection of structural damages [29]. The proposition emphasized cross-layer and cross-scale feature fusion to enhance the robustness of the detector. Kulkarni et al. proposed integrating principal component thermography analysis and EfficientDet [30] in detecting structural voids [31]. An extensive field test on a highway was conducted to validate the proposition. Also, DaCrack [32], an unsupervised domain adaptation framework for civil infrastructure crack detection, which utilizes contrastive mechanisms, adversarial learning techniques, and variational autoencoders to execute domain adaptation across the input, feature, and output dimensions, was proposed in a vision-based detection task. Similarly, a crack detection model that integrated an enhanced chicken swarm algorithm in a detection model was proposed for vision-based structural damage detection—the enhanced chicken swarm algorithm aimed at optimizing the detection model to enhance generalization capabilities [33]. Further, an end-to-end near-real-time detection model was proposed to detect structural anomalies (efflorescence) in brick walls. Likewise, an application-based deployment of state-of-the-art object detectors in detecting variant structural anomalies (efflorescence, corrosion, salient crack, debris, erosion) associated with civil infrastructure was presented [34]. In another application-based study [35], a lightweight inception and concatenation residual (InCR) detector, which leverages inception and concatenated residual blocks, was proposed for structural-level damage detection of buildings. Informative features were selected based on neighborhood component analysis before detecting damaged buildings. Ye et al. introduced the YOLOv7-AMF model, which consists of three distinct modules: Myswin and Aatten. These modules were designed to enhance the features of the model [36]. The third module presented in YOLOv7-AMF is the Feature Expansion and Enhancement Module (FEEM). It serves as a self-research module to augment the network's perceptual fields and enhance its overall performance. Additionally, Gao et al. proposed a framework for damage detection and volumetric assessment using LiDAR point clouds, integrating DeepLabV3+ for surface damage detection and a voxel-based approach for volumetric quantification [37]. The DeepLabV3+ model utilizes pseudo grayscale images derived from point cloud depth information, achieving a mean pixel accuracy of 97.2% for cracks exceeding 5mm in width and 6mm in depth. However, its performance diminishes for narrow cracks and under low point density or occlusion conditions. Also, the framework is constrained by its dependence on high-quality LiDAR data and limited effectiveness in detecting narrow cracks, highlighting areas for future improvement.

In summary, the current literature reveals the significant advancement of automated SHM via the integration of deep learning and varying computer-vision tasks. Within the computer vision tasks, the majority of SHM classification approaches leverage transfer learning, hybrid architectures, and probabilistic models to improve the accuracy of damage classification. The detection methods mostly emphasize robust model optimization, feature fusion, and real-time applications. Regardless of these advancements, challenges such as complex image environments and noisy data persist, necessitating further research to develop adaptive models for automated SHM.

*2.2. Near hierarchical-based structural damage detection and assessment via deep learning*

Hierarchical information associated with civil infrastructure data used to train object detectors in vision-based SHM applications has the potential to accelerate the performance and decision-making of object detectors. Not many studies have explored the hierarchical information associated with data in structural damage detection tasks.

A micro-drone-based multi-tasking framework, EnsembleDetNet, capable of multiple detection and scene-level classification, was proposed based on ensemble learning [38]. EnsembleDetNet used ensemble learning to induce diversity and strength correlation to enhance robustness. A recent similar study [39] proposed a framework, ExpoDet, characterized by scene and damage level assessment, automated surface damage aggregation scheme, and a weakly automated drone navigation module. The study employed two levels of related structural information, scene and damage level, in the vision-based SHM task but did not establish a link between the two. Also, Gao et al. [40] redefined vision-based damage classification to establish an intertask relationship between the abundant information related to structural damage images but did not utilize the hierarchical information; this establishes the Next Pacific Earthquake Engineering Research (PEER) Hub ImageNet dataset (Ø-NeXt). A TransUnet was also proposed to extract multi-scale damage features related to concrete structures prior to detection [41]. The Transformer block was employed to improve the self-attention module in the TransUNet to capture multiple contextual and global knowledge prior to the detection of damages. Lastly, a damage detection and classification model centered on high-level detection and low-level classification tasks was proposed [42]. The objective of the high-level detection task is to distinguish images that exhibit damage from undamaged images. Afterward, the low-level classification task computes the likelihood of each damage under the assumption that the image contains defects. The overall classification of defects is subsequently derived using the chain rule of conditional probability.

From the current reiterated literature, apart from the work of Gao et al. [40], the closely related intertask hierarchical vision-based SHM studies have mostly focused on multi-tasking SHM without linkage between the tasks as well as the use of the hierarchical relationship between the tasks. This leaves out the hierarchical information associated with structural damages in civil infrastructure; this abundant hierarchical intertask information can potentially improve the performance of varying DCNN-based models in vision-based SHM and other vision-based applications. To this end, as a contribution, this study proposes a Guided-DetNet framework. The framework demonstrates how hierarchical intertask information associated with structural damages can improve the performance of DCNN-based models in vision-based SHM applications, thereby establishing a linkage between multiple variant tasks and utilizing the hierarchical information to refine class instances.

## 3. Proposed Guided-DetNet framework
*3.1. Overview*

Guided-DetNet addresses three challenges: (1) complex image environments, (2) noisy labeling, and (3) surface damage assessments post-detections via (1) generative attention module, (2) using hierarchical structural damage information, and (3) volumetric contour visual assessment, respectively. Guided-DetNet is tailored for two levels of SHM vision-based tasks: triple classification and dual detection tasks, respectively. The Guided-DetNet framework comprises a multi-classification classifier (modified YOLOv8-m herein YOLOv8-m-GAM-Net) with three classification heads, which guides the detector (modified YOLOv8-m herein YOLOv8-m-GAM-Det) with dual decoupled heads for multi-detection tasks. The classifier and the detector cooperate in a vision-based task via the proposed hierarchical elimination algorithm to mitigate noisy labeling and enhance performance. Additionally, to deal with complex image environments, the proposed generative attention module, which generates additional multi-varied feature maps, is embedded in both DCNN models in the Guided-DetNet to enhance the percepts of the models. Lastly, the detection of damages is assessed via a volumetric contour visual assessments module within the Guided-DetNet, which quantifies the extent of structural visual damages. Fig. 1 illustrates the proposed Guided-DetNet framework, while the subsequent sections under section 3 delve deeper into the various modules of Guided-DetNet.

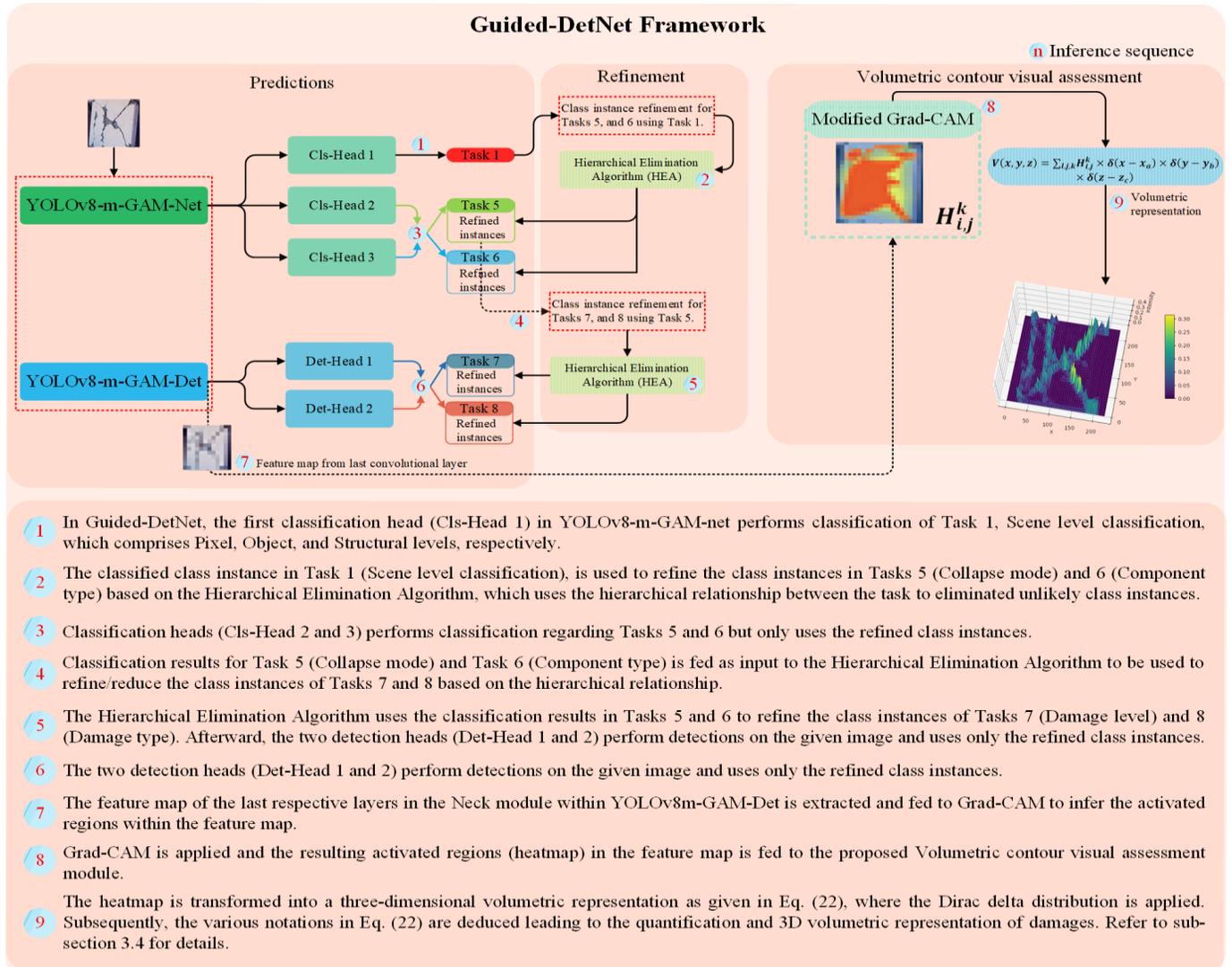

**Fig. 1.** Overview of Guided-DetNet framework for triple classification, dual detection, and volumetric quantification of detected damages.

## 3.2. Generative attention module (GAM)

The proposed Generative Attention Module (GAM), with characteristics of multi-varied feature map generation, is an advancement over our two previous attention modules; thus, the explicit ensemble attention module (EEAM) [38] and explicit ensemble attention module plus (EEAM+) [39].

EEAM (see Fig. 2a) and EEAM+ (see Fig. 2b) leveraged convolutional block attention modules (CBAM) [43] and parallel attention modules (PAM) [44]; both CBAM and PAM utilize spatial attention modules (SAM) and channel attention modules (CAM), which address where to search and what features to search for, respectively. To enhance features by EEAM, an explicit ensembling of average pooled features $w_x \boldsymbol{F}''_{avg}, w_y \boldsymbol{P}''_{avg}$ and max-pooled features $z_x \boldsymbol{F}''_{max}, z_y \boldsymbol{P}''_{max}$ from CBAM and PAM, respectively, via a convolutional operation takes place. In the average pooled features, $w_x$ and $w_y$ are tuning hyper-parameters and $z_x$ and $z_y$ are weight adjustments in the max-pooled features. Further, a concatenation operation of the explicit ensembled average pooled features $\boldsymbol{E}_{avg}$ and explicit ensembled max-pooled features $\boldsymbol{E}_{max}$ occur, leading to a refined feature map $\boldsymbol{R}^{F'' \oplus P''}$, where $\boldsymbol{F}''$ and $\boldsymbol{P}''$ are feature descriptors from CBAM and PAM. Also, EEAM+, an improvement over EEAM, addresses how to search for features by introducing dynamic rotation of feature maps. In EEAM+, refined feature map plus are produced by first taking in a feature map $\boldsymbol{F}_{map}$, afterward the feature map $\boldsymbol{F}_{map}$ is transposed by a direction $\vec{D}$ and an angle $\theta$. The angle $\theta \in \mathbb{A}$ where $\mathbb{A} = \pm\{90, 180, 270\}$, and the direction $\vec{D}$ is either clockwise or anti-clockwise. After the transpose operation, the feature maps go through the explicit ensembling and concatenation operation as summarized in the EEAM; this results in the refined feature map plus $\boldsymbol{R}^{F'' \oplus P''}_{plus}$. Although EEAM and EEAM+ improved the performance of the frameworks in our previous studies, the two mentioned attention modules are limited in terms of extra diversified feature maps. EEAM and EEAM+ take in a feature map $\boldsymbol{F}_{map}$, and produces an enhanced feature map and a dynamic enhanced feature map, respectively, without increasing the depth (additional feature maps) of the feature maps to increase diversity.

To address the limitations of EEAM and EEAM+, the generative attention module (GAM), which goes beyond feature enhancement but can generate multi-varied feature maps to increase the diversity of feature maps, is proposed. Unlike standard attention mechanisms, GAM leverages EEAM and EEAM+ to generate multi-varied feature maps via two main sub-modules: (1) cross-horizontal and cross-vertical patch merging and (2) cross foreground-background feature fusion; this enhances robustness in complex environments by diversifying feature representations.

### 3.2.1 Cross-horizontal and cross-vertical patch merging

Given a refined feature map $\boldsymbol{R}^{F'' \oplus P''}$ and a refined feature map plus $\boldsymbol{R}^{F'' \oplus P''}_{plus}$ from EEAM and EEAM+, respectively, where $\boldsymbol{R}^{F'' \oplus P''}$ and $\boldsymbol{R}^{F'' \oplus P''}_{plus} \in \mathbb{R}^{C \times W \times H}$ ($C, W,$ and $H$ represent a two-dimensional RGB image channel, width, and height, respectively); two types of patch merging are used in generating new multi-varied feature maps, as seen in Fig. 2c. The feature maps $\boldsymbol{R}^{F'' \oplus P''}$ and $\boldsymbol{R}^{F'' \oplus P''}_{plus}$ can be expressed as Eq. 1:

$$\boldsymbol{R}^{F'' \oplus P''}, \boldsymbol{R}^{F'' \oplus P''}_{plus} = \begin{bmatrix} P_1^{r1}, P_2^{r1}, \cdots, P_n^{r1} \\ P_1^{r2}, P_2^{r2}, \cdots, P_n^{r2} \\ \vdots & \vdots & \ddots & \vdots \\ P_1^{rn}, P_2^{rn}, \cdots, P_n^{rn} \end{bmatrix} \quad (1)$$

where $P_n^{rn}$ represent the pixel index of the feature map, $rn$, and $n$ denote their rows and columns, respectively. To generate multi-varied features, the given feature maps $\boldsymbol{R}^{F'' \oplus P''}$ and $\boldsymbol{R}^{F'' \oplus P''}_{plus}$ are partitioned into two patches (horizontal and vertical, see Fig. 2c) and indexed 1 and 2 as expressed in Eq. 2 and Eq.3 (In Eq. 3, letters *A*, and *B* are used only for easier understanding in the mathematical representation).

$$\boldsymbol{R}^{F'' \oplus P''} = \begin{bmatrix} Patch\ 1 \\ Patch\ 2 \end{bmatrix} and\ [Patch\ 1 \quad Patch\ 2] \quad (2)$$

$$\boldsymbol{R}^{F'' \oplus P''}_{plus} = \begin{bmatrix} Patch\ A \\ Patch\ B \end{bmatrix} and\ [Patch\ A \quad Patch\ B] \quad (3)$$

Given feature maps ($\boldsymbol{R}^{F'' \oplus P''}$ and $\boldsymbol{R}^{F'' \oplus P''}_{plus}$), with even dimensions (e.g., 224 × 224), to patch the feature maps into two (horizontal and vertical), the width $W$, or height $H$ of the feature maps ($\boldsymbol{R}^{F'' \oplus P''}$ and $\boldsymbol{R}^{F'' \oplus P''}_{plus}$) are first divided by 2 to derive the width $W_{patch}$, and height $H_{patch}$ of the patches (i.e., the column split index $ClP_n^{rn}$ and row split index $RwP_n^{rn}$ of the feature maps, respectively) as represented in Eq. 4:

$$W_{patch}, H_{patch} = \frac{W}{2}, \frac{H}{2} \quad (4)$$

For odd-dimension feature maps ($R^{F'' \oplus P''}$ and $R_{plus}^{F'' \oplus P''}$), to derive the width $W_{patch}$, and height $H_{patch}$ of the patches, the floor and ceiling functions are utilized to round down and round up the output after the division operation since odd dimensions are not easily divisible by 2.

As a practical example, if the dimensions of the feature maps ($R^{F'' \oplus P''}$ and $R_{plus}^{F'' \oplus P''}$) are $127 \times 127$ (width $W$, and height $H$), the floor function $\lfloor x \rfloor$ ($x$ being a real number) is applied to the refined feature map $R^{F'' \oplus P''}$ while the ceiling function $\lceil y \rceil$ ($y$ being a real number) is applied to the refined feature map plus $R_{plus}^{F'' \oplus P''}$; this results in the row and column split indexes ($RwP_n^{rn}$ and $ClP_n^{rn}$), respectively, as given in Eq. (5) and Eq. (6) for horizontal split and Eq. (7) and Eq. (8) for vertical split.

$$RwP_n^{rn} \equiv H_{patch}\left(R^{F'' \oplus P''}\right) = \left\lfloor \frac{H}{2} \right\rfloor = \left\lfloor \frac{127}{2} \right\rfloor = 63 \quad (5)$$

$$RwP_n^{rn} \equiv H_{patch}\left(R_{plus}^{F'' \oplus P''}\right) = \left\lceil \frac{H}{2} \right\rceil = \left\lceil \frac{127}{2} \right\rceil = 64 \quad (6)$$

$$ClP_n^{rn} \equiv W_{patch}\left(R^{F'' \oplus P''}\right) = \left\lfloor \frac{W}{2} \right\rfloor = \left\lfloor \frac{127}{2} \right\rfloor = 63 \quad (7)$$

$$ClP_n^{rn} \equiv W_{patch}\left(R_{plus}^{F'' \oplus P''}\right) = \left\lceil \frac{W}{2} \right\rceil = \left\lceil \frac{127}{2} \right\rceil = 64 \quad (8)$$

The results of Eq. (5) to Eq. (8) lead to the dimensions of two patches for each feature map ($R^{F'' \oplus P''}$ and $R_{plus}^{F'' \oplus P''}$) with one patch being slightly smaller than the other. Eqs. (9) and (10) represent the horizontal split of the feature maps and Eqs. (11) and (12) represent the vertical split of the feature maps.

$$R^{F'' \oplus P''} = \begin{bmatrix} Patch\ 1(W \times H) = (127 \times 63) \\ Patch\ 2(W \times H) = (127 \times 64) \end{bmatrix} \quad (9)$$

$$R_{plus}^{F'' \oplus P''} = \begin{bmatrix} Patch\ A(W \times H) = (127 \times 64) \\ Patch\ B(W \times H) = (127 \times 63) \end{bmatrix} \quad (10)$$

$$R^{F'' \oplus P''} = [Patch\ 1\ (W \times H) \quad Patch\ 2\ (W \times H) \\ = (63 \times 127), (64 \times 127)] \quad (11)$$

$$R_{plus}^{F'' \oplus P''} = [Patch\ A\ (W \times H) \quad Patch\ B\ (W \times H) \\ = (64 \times 127), (63 \times 127)] \quad (12)$$

After performing the patching operation for even dimensions, as given in Eq. 4, and odd dimensions, as given in Eqs. 5 to Eq. 8, a merging operation via (1) cross horizontal merge and (2) cross vertical merge occurs. Since even-sized dimension patches can be merged easily, the odd-sized dimension is used for additional explanation. Given two horizontal patched feature maps as represented in Eqs. (9) and (10), the cross-horizontal merging results in two new multi-varied feature maps $mvF_{map}^n$, where $n$ denotes the index of the generated multi-varied feature map, as in Eqs. (13) and (14).

$$mvF_{map}^1 = \begin{bmatrix} Patch\ 1\ (W \times H) = (127 \times 63) \\ Patch\ A\ (W \times H) = (127 \times 64) \end{bmatrix} \quad (13)$$

$$mvF_{map}^2 = \begin{bmatrix} Patch\ 2\ (W \times H) = (127 \times 64) \\ Patch\ B\ (W \times H) = (127 \times 63) \end{bmatrix} \quad (14)$$

Similarly, in a cross-vertical merging operation of Eqs. (11) and (12), two more multi-varied feature maps are generated as given in Eq.s (15) and (16).

$$mvF_{map}^3 = \begin{bmatrix} Patch\ 1\ (W \times H) = (63 \times 127) \\ Patch\ A\ (W \times H) = (64 \times 127) \end{bmatrix} \quad (15)$$

$$mvF_{map}^4 = \begin{bmatrix} Patch\ 2\ (W \times H) = (63 \times 127) \\ Patch\ B\ (W \times H) = (64 \times 127) \end{bmatrix} \quad (16)$$

Fig. 2e gives a graphical insight into the patching and merging operation for odd and even-sized dimensions of feature maps.

Overall, the cross-horizontal and cross-vertical patching and merging operation leads to four newly generated multi-varied feature maps ($mvF_{map}^1, \ldots, mvF_{map}^4$) based on two feature maps ($R^{F'' \oplus P''}$ and $R_{plus}^{F'' \oplus P''}$), which are from the EEAM and EEAM+, respectively.

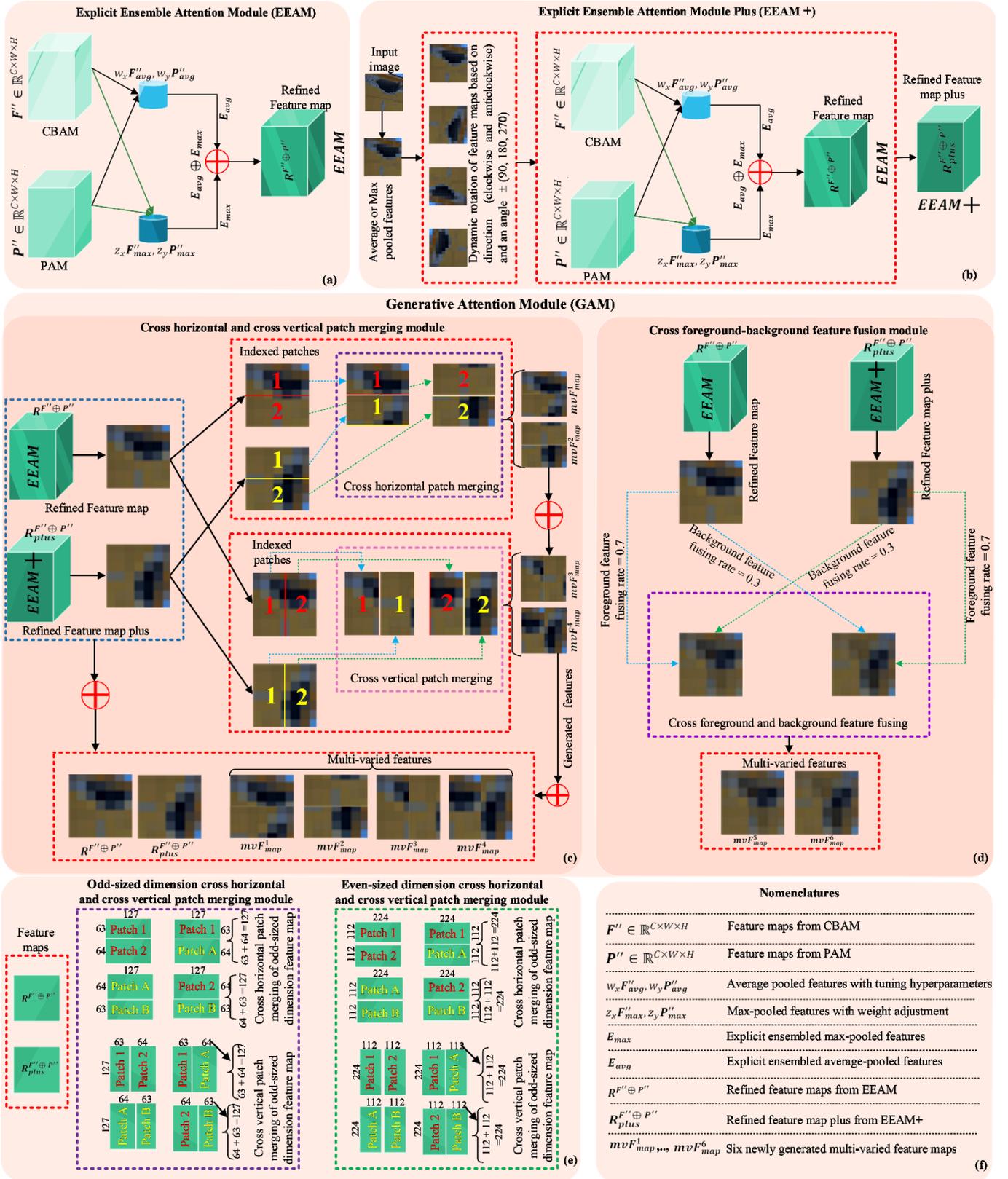

**Fig. 2.** Graphical representation of the proposed Generative attention module (GAM): (a) Explicit Ensemble Attention Module (EEAM), (b) Explicit Ensemble Attention Module Plus (EEAM+), (c) illustration of the proposed cross horizontal and cross vertical patch merging sub-module in GAM, (d) illustration of the proposed cross-foreground-background feature fusion sub-module in GAM, (e) representation of patch and merge operation for odd and even-sized dimensions, and (d) explanation of nomenclatures in Fig. 2.

*3.2.2 Cross foreground-background feature fusion*

In addition to the cross-horizontal and cross-vertical patch merging sub-module in the GAM, which generates four new multi-varied feature maps ($mvF_{map}^1,\ldots, mvF_{map}^4$); two more additional multi-varied feature maps ($mvF_{map}^5$, and $mvF_{map}^6$) are generated based on the feature maps ($R^{F''\oplus P''}$ and $R_{plus}^{F''\oplus P''}$) from EEAM and EEAM+ via a cross-foreground-background feature fusion module, as seen in Fig. 2d.

In the cross-foreground-background feature fusion module, the goal is to emphasize the features present in the refined feature map $R^{F''\oplus P''}$ while incorporating some aspects of the refined feature map plus $R_{plus}^{F''\oplus P''}$ and vice-versa based on a fusing rate $F_{rate}$. First, to generate the fifth multi-varied feature map $mvF_{map}^5$, the fusion operation is expressed as Eq. (17).

$$mvF_{map}^5(x,y) = 1-(F_{rate}) \times R^{F''\oplus P''}(x,y) + F_{rate} \times R_{plus}^{F''\oplus P''}(x,y) \quad (17)$$

In $mvF_{map}^5(x,y)$, $(x,y)$ represents the pixel value at position $(x,y)$ in the crossed foreground-background fused feature map. Also, $(x,y)$ in the feature maps ($R^{F''\oplus P''}(x,y)$ and $R_{plus}^{F''\oplus P''}(x,y)$) are the pixel values of the refined feature map $R^{F''\oplus P''}$, and the refined feature map plus $R_{plus}^{F''\oplus P''}$, respectively, and the fusing rate $F_{rate}$ is set at 0.7. The result of Eq. (17) is a crossed foreground-background feature map where the refined feature map $R^{F''\oplus P''}$ dominates in the foreground, and the refined feature map plus $R_{plus}^{F''\oplus P''}$ contributes to the background in the final appearance based on the specified fusing rate.

Similarly, to generate the sixth multi-varied feature map $mvF_{map}^6$, the refined feature map plus $R_{plus}^{F''\oplus P''}$ dominates in the foreground, while the refined feature map $R^{F''\oplus P''}$ serves in the background. The fusion operation is given in Eq. (18)

$$mvF_{map}^6(x,y) = F_{rate} \times R^{F''\oplus P''}(x,y) + (1-F_{rate}) \times R_{plus}^{F''\oplus P''}(x,y) \quad (18)$$

For brevity, $(x,y)$ in Eq. (18) is explained as in Eq. (17). Also, the fusion rate $F_{rate}$ is set as 0.7. These cross-foreground-background fusion methods provide flexibility in controlling the feature dominance of each feature map ($R^{F''\oplus P''}$ and $R_{plus}^{F''\oplus P''}$) from EEAM and EEAM+ in generating multi-varied feature maps ($mvF_{map}^5$, and $mvF_{map}^6$).

Overall, from Fig 2, combining the feature maps generated based on the two sub-modules ((1) cross-horizontal and cross-vertical patch merging and (2) cross foreground-background feature fusion) in GAM, six new multi-varied feature maps ($mvF_{map}^1,\ldots, mvF_{map}^6$) are generated to increase the Guided-DetNet framework's diversity, further strengthening its robustness.

*3.3 GAM-YOLOv8*

You Only Look Once version eight (YOLOv8) [45], the current version of the YOLO series, as of 2023, is a cutting-edge object detection model designed for object detection, instance segmentation, and image classification. The architecture, which has three main modules (backbone, neck, and head), integrates advanced features to enhance accuracy and speed in detecting objects within images.

The backbone of YOLOv8 leverages the Cross Stage Partial Darknet53 (CSPDarknet53) [46] feature extractor, similar to one of its predecessors, the YOLOv5 [47], with a notable enhancement – including the C2f module. This cross-stage partial bottleneck incorporates two convolutions, fusing high-level features with contextual information to enhance detection accuracy. Also, the C2f has the benefit of ensuring the flow of abundant gradient information. In addition to these features of the YOLOv8, the proposed GAM in this study is embedded in the backbone to enhance the feature descriptors and increase the feature maps' depth post-convolutional operations; this ensures significant diversity of feature maps, leading to the robustness of the framework.

The neck module of the YOLOv8 adopts a blend between the Path Aggregation Network (PANet) [48] and Feature Pyramid Network (FPN) [49]. As such, the neck of YOLOv8 has the advantage of PANet, which increases information flow using the bottom-up path augmentation concept. This aids in shortening information flow between the lower and top layers, respectively, contributing to the feature hierarchy of localization with precision. Also, the use of top-down lateral connections to exploit multi-scale pyramid hierarchy to yield high-level feature maps at varying scales, which is the advantage of the FPN, is experienced in the neck of YOLOv8. As an addition to the neck structure of YOLOv8, GAM is embedded before PANet and FPN; this allows both sub-modules to exploit the abundant multi-varied feature maps ($mvF_{map}^1,\ldots, mvF_{map}^6$) generated by GAM to increase robustness, leading to improved performance.

In the head, YOLOv8 adopts a decoupled structure similar to YOLOX [50] head architecture, separating the classification and detection heads. This significant re-engineering of the decoupled head in YOLOv8 is complemented by a shift from Anchor-Based to Anchor-Free methodology in positioning objects through the center, predicting the distance between the center and the bounding box. The output layer of YOLOv8 utilizes a sigmoid activation function, producing probability for detected objects.

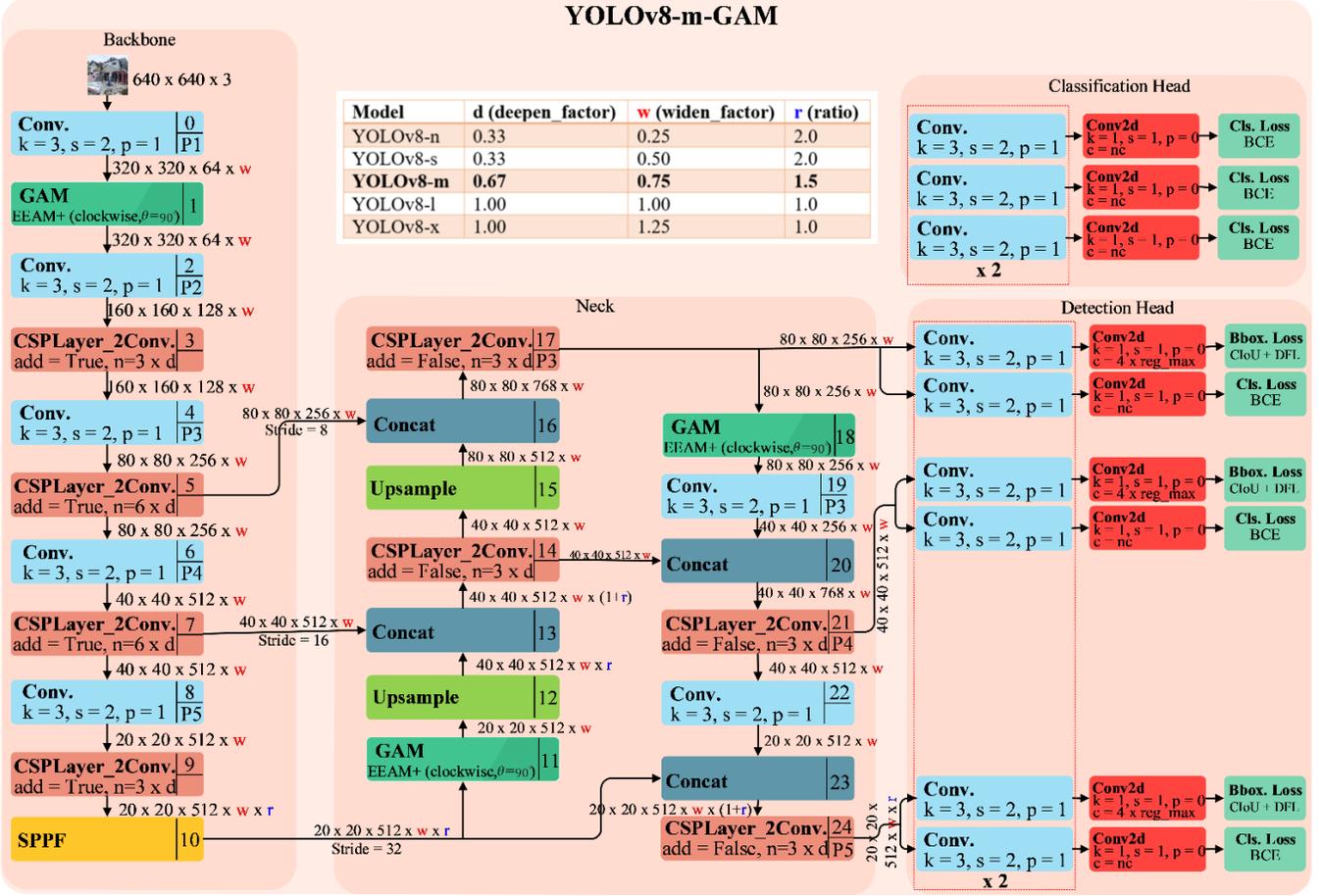

**Fig. 3.** Graphical representation of the base model, YOLOv8, embedded with the proposed Generative Attention Module (GAM).

Notably, a TaskAlignedAssigner, which assigns labels to anchor boxes, and a Distribution Focal Loss (DFL) tailored for classification in the context of object detection with a focus on addressing class imbalance issues are the modifications in the output layer. These modifications contribute to higher detection accuracy and faster processing speeds compared to previous YOLO versions.

In addition to the YOLOv8's innovative architectural features (C2f module, decoupled head, and Anchor-Free methodology), the proposed GAM is embedded strategically to increase the depth of the feature maps to increase robustness and improve accuracy. Since the EEAM+ in GAM uses varying direction $\vec{D}$ (clockwise or anti-clockwise) and rotational angle $\theta$, in Fig. 3, the specified direction $\vec{D}$ and rotational angle $\theta$ are given.

### 3.4 Hierarchical elimination algorithm and Guided-DetNet framework

In the pursuit of developing an effective multi-task framework for structural damage detection, an essential consideration is a hierarchical relationship between the intertask (e.g., scene level, damage level, and others), which can help reduce noisy labels that undermine the framework's performances. The PEER Hub Image-Net (Ø-Net) [51] dataset, which comprises eight intertasks as visualized in Fig. 4, is utilized in this study, focusing on Tasks 1, 5, 6, 7, and 8, respectively.

From Fig. 4 and the structural damage engineering description and judgment in Table 1, Task 1 (scene level) is the easiest, with arguably no noisy labels since annotators can easily differentiate between pixel, object, and structural levels of infrastructure-related images. As a result, the hierarchical information of Task 1 is used as a guide for inference by eliminating unlikely class instances on a moderately challenging task; thus, Tasks 5 (collapse mode) and 6 (component type), respectively. Subsequently, the hierarchical relationship of Tasks 5 is used to guide the inference related to a more complex task: Tasks 7 (damage level) and 8 (damage type), respectively. The hierarchical relationship provides a significant understanding of the intertask of Ø-Net data. It forms the basis for the proposed Hierarchical Elimination Algorithm (HEA), which eliminates some class instances (most unlikely class instance of a Task) during inference based on the hierarchical relationship of other Tasks; this strengthens the framework's prediction accuracy. To be more practical, given an image $I$, which has associated varied class instances $C^n$ as in Table 1 for Task $T^n \in \mathbb{T}$, where $\mathbb{T} = (Task\ 1, Task\ 5, Task\ 6, Task\ 7, Task\ 8)$, if the class instance prediction for image $I$ in

**Algorithm 1:** Hierarchical Elimination Algorithm

Function refine_class_instances(task1_prediction, task5_prediction):

1.     $T^{Task\,5}\_C^n$ = ["None", "Partial", "Global"]
2.     $T^{Task\,6}\_C^n$ = ["Beam", "Column", "Wall", "Other"]
3.     *# Elimination mechanism for Task 5 and Task 6*
4.     **if** task1_prediction = $C^{Pixel}$ or $C^{Object}$:
5.         $T^{Task\,5}\_C^n = T^{Task\,5}\_C^n$ – ["Global"]
6.         $T^{Task\,6}\_C^n = T^{Task\,6}\_C^n$ – ["Other"]
7.     **else**:
8.         $T^{Task\,5}\_C^n = T^{Task\,5}\_C^n$ – ["Partial"]
9.         $T^{Task\,6}\_C^n = T^{Task\,6}\_C^n$ – ["Beam", "Column", "Wall"]
10.     *# refined class instances of Tasks 5 & 6*
11.     $C^{Task\,5^n}, C^{Task\,6^n} = [T^{Task\,5}\_C^n, T^{Task\,5}\_C^n]$
12.     $T^{Task\,7}\_C^n$ = ["Undamaged", "Minor", "Moderate", "Heavy"]
13.     $T^{Task\,8}\_C^n$ = ["Undamaged", "Flexural", "Shear", "Combined"]
14.     *# Elimination mechanism for Task 7 and Task 8*
15.     **if** task5_prediction = $C^{None}$:
16.         $T^{Task\,7}\_C^n = T^{Task\,7}\_C^n$ – ["Moderate", "Heavy"]
17.     **else if** task5_prediction = $C^{Partial}$:
18.         $T^{Task\,7}\_C^n = T^{Task\,7}\_C^n$ – ["Undamaged", "Heavy"]
19.         $T^{Task\,8}\_C^n = T^{Task\,8}\_C^n$ – ["Undamaged"]
20.     **else**:
21.         $T^{Task\,7}\_C^n = T^{Task\,7}\_C^n$ – ["Undamaged", "Minor"]
22.         $T^{Task\,8}\_C^n = T^{Task\,8}\_C^n$ – ["Undamaged", "Minor"]
23.     *# refined class instances of Tasks 7 & 8*
24.     $C^{Task\,7^n}, C^{Task\,8^n} = [T^{Task\,7}\_C^n, T^{Task\,8}\_C^n]$
25.     **return** $C^{Task\,5^n}, C^{Task\,6^n}, C^{Task\,7^n}, C^{Task\,8^n}$

$T^{Task\,1}$ is object $C^{Object}$, it can be used to eliminate some class instances in Task $T^{Task\,5}$ to derive a refined class instance $C^{*n}$ (* represents $T^n$) prior to the prediction of the given image $I$ for $T^{Task\,5}$ as follows:

$$C^{Task\,5^n} = T^{Task\,5} \frac{\{C^{None}, C^{Partial}, C^{Global}\}}{\{C^{Object}\}} \tag{19}$$

$$C^{Task\,5^n} = T^{Task\,5} \{C^{None}, C^{Partial}\} \tag{20}$$

From Eq. 19, class prediction $C^{Object}$ of $T^{Task\,1}$ is used to eliminate the class instance $C^{Global}$ of $T^{Task\,5}$ based on the hierarchical relationship since the global $C^{Global}$ refers to the entire infrastructure state of collapse assessment and object $C^{Object}$ refers to the scene-level understanding of the infrastructure. A detailed hierarchical relationship of Task 1 class instances to Tasks 5 and 6 and Task 5 class instances to Tasks 7 and 8 is given in sub-section 3.3.1. The HEA, which refines the class instances for a particular Task $T^n$ is given in algorithm 1.

*3.4.1 Hierarchical relationship between classes of Tasks*

First, the study considers two levels of SHM vision-based tasks: (1) a triple classification task and (2) a dual detection task. As a result, the classifier (YOLOv8-GAM-Net) and the detector (YOLOv8-GAM-Det) in Guided-DetNet have a triple decoupled head and a dual decoupled head, respectively, as seen in Fig. 1.

In the triple classification task, the relationship between Task 1 (scene level) is used to guide the classification of class instances in Task 5 (collapse mode) and Task 6 (component type). Task 1's pixel level (P) involves close-up views, which does not give much information about a structural part; hence, in the context of Task 5, an image classified as pixel level in Task 1 cannot be categorized as partial (PC) or global (GC) collapse in Task 5; hence global and partial class instances are eliminated. Also, Task 1's object (O) level corresponds to partial views (infrastructure parts such as beam, pillar, and others); this has a strong relationship with Task 5's none (N) class instance and partial (PC) class instance, which in the presence of damage, signifies that only part of the structure has collapsed while other parts remain intact. As such, Task 5's global class instance is eliminated in this case. The structural level (S) in Task 1 encompasses most or all parts of an infrastructure; hence, in Task 5, this information aids in the classification of global collapse (GC) in the presence of damage, which indicates severe damage where most or the entire infrastructure has collapsed. Also, in the absence of damage or the presence of thin defects, Task 1's structural level classification complements the none (N) class instance in Task 5; hence, the partial (PC) class instance is eliminated in Task 5. Similarly, Task 1's classification information aids in Task 6 classification by eliminating some class instances of Task 6. Task 6 focuses on structural parts (beam, wall, column, others), in which pixel-level scene classification reveals partial information leading to the structural description of beams, walls, columns, and others. As a result, Task 1's pixel- and object-level class instances do not eliminate any class instances in Task 6 since pixel- and object-level information reveals information leading to the classification of any instances in Task 6. Since the structural level of Task 1 reveals an entire structure comprising mostly one or more structural parts (beam, wall, column), it is challenging to classify a component type (Task 6) at the structural level. As a result, Task 1's structural level class instance is used to eliminate beam, column, and wall, leaving the other class instance in Task 6, which means an unidentified distinct structural part or a collection of several structural parts.

**Table 1**
The description of tasks and class instances of the tasks.

| Task $T^n$ | Classes $C^n$ | Task and class instances description |
|---|---|---|
| Task 1 (scene level) | Pixel (P), Object (O), Structural (S) | Task 1 classifies images into pixel level (P), object level (O), or structural level (S) to provide a foundational understanding of the scene. The pixel level entails partial information on a structural part, and the object level reveals a complete structural part, while the structural level reveals an entire infrastructure. |
| Task 5 (collapse mode) | None (NC), Partial (PC), Global (GC)_ | Task 5 evaluates and categorizes damage severity into collapse modes, including none (NC), partial (PC), or global (GC). None (NC) corresponds to no structural collapse but may slightly surface damage patterns, and partial (PC) means a collapse of part of an infrastructure while other parts remain intact. Global (GC) means a complete collapse of infrastructure. |
| Task 6 (compoent type) | Beam (Bm), Column (Cm), Wall (Wl), Other (Or) | Task 6 identifies structural components and classifies them into beams (Bm), columns (Cm), walls (Wl), or others (Or). Beams represent horizontal structural parts that support and transfer loads horizontally to the columns, which are vertical structural parts that transfer the received loads to the foundation of the infrastructure. Walls are vertical load-bearing structures that resist lateral forces and provide stability to an infrastructure. Other represents an unidentified structural part of an infrastructure or a collection of several structural parts (beam, column, wall). |
| Task 7 (damage level) | Undamaged (UD), Minor (MiD), Moderate (MoD), Heavy | Task 7 assesses damage and categorizes it into undamaged (UD), minor damage (MiD), moderate damage (MoD), or heavy damage (HvD). Undamaged means the absence of damage, and minor damage represents slight patterns of thin cracks. Moderate damage indicates the minor damage level has progressed to encompass moderate parts of the infrastructure, and heavy damage indicates a part or the entire infrastructure is nearing failure or has collapsed. |
| Task 8 (damage type) | Undamaged (UnD), Flexural (Flex), Shear, Combined (Comb) | Task 8 defines complex damage patterns as undamaged (UnD), flexural (FLEX), shear (SHEAR), or combined (COMB). Undamaged means a complete absence of damages, and flexural represents structural damages (molds, efflorescence, scaling, cracks, curling) occurring in horizontal or/and vertical patterns. Shear represents structural damage (cracks, spalling, curling) patterns of complete or partial shapes in the form of X, Y, V, or diagonal patterns. Combined damage refers to structural damage that exhibits an uneven pattern, in contrast to flexural and shear damage. |

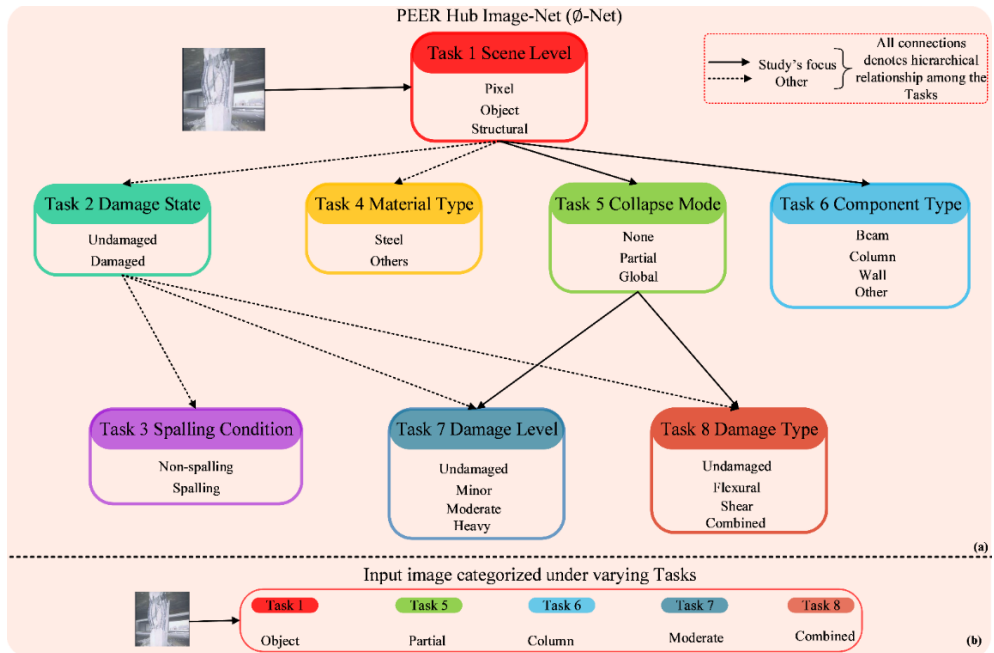

**Fig. 4.** Illustration of the relationship between inter-tasks of the PEER Hub-ImageNet.

In the dual detection tasks, classification information relating to Task 5 is used to eliminate the most unlikely class instances in Tasks 7 and 8, respectively. None (N) class instance in Task 5 corresponds to undamaged or slightly damaged patterns of an infrastructure. As a result, when an image is classified as none in Task 5, it is unlikely for the image to be either moderately or heavily damaged in Task 7; hence, this information is used to eliminate moderate and heavy class instances. Also, if an image is classified as partially collapsed, then it is most likely the image will be either minor or moderately damaged in Task 7; hence, undamaged and heavy class instances are eliminated in this scenario. Lastly, once an image is classified as global in Task 5, the undamaged and minor class instances are eliminated since it is unlikely for an infrastructure to be entirely collapsed and be categorized as either undamaged or minor regarding damage levels. Similarly, predictions in Task 5 are utilized to eliminate some class instances in Task 8 (damage type), which have undamaged, flexural, shear, and combined as class instances. Once an image is classified as partial or global in Task 5, the undamaged class in Task 8 is eliminated since a partial or global description of a collapsed infrastructure cannot be termed as undamaged in terms of damage type. Lastly, if the image is classified as none regarding collapse mode in Task 5, no class instance in Task 8 is eliminated since such an image has the possibility of belonging to any of the damage types of Task 8.

Overall, the hierarchical information from Task 1 and Task 5 provides a valuable guide and ensures a coherent evaluation of a given image under the subsequent Tasks 5, 6, 7, and 8, respectively.

*3.5 Volumetric contour visual assessment*

The assessment of damages in critical infrastructure after damage detection is essential for effective decision-making and maintenance. Gustavo et al. proposed volumetric quantification of damages post-detection via the use of an RGB-D sensor; however, critical aspects, such as the representation of the quantified damages in 3D, were not considered [52]. Additionally, not many existing detection models or frameworks take into consideration the assessments of detected damages. In order to mitigate this challenge, a three-dimensional (3D) Volumetric Contour Visual Assessment (VCVA) is proposed, utilizing an RGB. VCVA enables a richer understanding of the spatial distribution and severity of damages in 3D; it quantifies and visualizes the damages.

The proposed VCVA employs a modified Gradient-weighted Class Activation Mapping (Grad-CAM) to visualize the activated regions of an image. Given the output (either triple classification or dual detection task) of Guided-DetNet, for an input image $I$, the modified Grad-CAM generates a heatmap $H_{i,j}^k$ for each feature map of index $k$ with spatial location $(i, j)$. The heatmap $H_{i,j}^k$, which is given in Eq. (21), highlights the focus area of Guided-DetNet, mostly regions in the image $I$ relating to the object of interest (class instances where the interest is solely on damages); this serves as a foundation for the subsequent volumetric analysis.

$$H_{i,j}^k = ReLU\left(\sum_l a_i^k \times A_{i,j}^{k,l}\right) \quad (21)$$

In Eq. (21), $l$ is the index of the channel in the feature map, $a_i^k$ is the weight assigned to each channel, $k$ represents the index of the feature map, $i, j$ denotes the spatial coordinates of the feature map, and $A^k$ represents the feature map at index $k$. The heatmap $H_{i,j}^k$ is transformed into a three-dimensional (3D) volumetric representation $V(x, y, z)$, as in Eq. (22), utilizing the Dirac delta distribution $\delta$.

$$V(x, y, z) = \sum_{l,j,k} H_{i,j}^k \times \delta(x - x_a) \times \delta(y - y_b) \times \delta(z - z_c) \quad (22)$$

In Eq. (22) $x, y, z$ is determined by discretizing the 3D space into a grid (set of coordinates $x_a, y_b, z_c$) evenly spaced, which represent the discrete spatial locations within a grid, and they correspond to the indices of the grid points. The values of $x, y, z$ correspond to the coordinates $(x_a, y_b, z_c)$ of a given grid point $a, b, c,$ multiplied by a grid spacing value $G_{space}$ (interval between adjacent grid points along $x, y, z$ axes); this is computed as follows:

$$x = a \times G_{space} \quad (23)$$

$$y = b \times G_{space} \quad (24)$$

$$z = c \times G_{space} \quad (25)$$

The Dirac delta distribution $\delta$ in Eq. (22), which is expressed as Eq. (26), is used to concentrate the information at specific spatial coordinates. In practice, the Dirac delta distribution $\delta$ is approximated with a finite value at a grid point closest to the true spatial location.

$$\delta(x - x_i) \approx \begin{cases} 1 & \text{if } |x - x_i| < \epsilon \\ 0 & \text{otherwise} \end{cases} \quad (26)$$

In Eq. (26), $\epsilon$ is $a_i^k$ (weights assigned to the channel of a feature map) from Eq. (21); this defines the peak of the Dirac delta function. Lastly, the volumetric representation $V(x, y, z)$ is quantified as Eq. (27) to assess the severity and extent of damages in the given image $I$.

$$V_{total} = \sum_{x,y,z} V(x, y, z) \quad (27)$$

Overall, the volumetric representation $V(x, y, z)$ encapsulates the aggregated information from the modified Grad-CAM heatmap $H_{i,j}^k$ across a feature map $A^k$ and discretized spatial coordinates. The discretization of the 3D space into a grid enables us to quantify the extent and severity of damages on the infrastructure.

## 4. Experiment

The study experimental section provides details regarding the dataset, configuration, and implementation, as well as the evaluation metrics used in assessing Guided-DetNet and the compared models.

### 4.1 PEER Hub Image-Net (∅-Net) dataset

The Pacific Earthquake Engineering Research (PEER) Hub Image-Net (∅-Net), which establishes a hierarchy among varying tasks, is used in this study. The dataset has eight levels of inter-task, out of which the study focuses on five tasks; thus Task 1 (scene level), Task 5 (collapse mode), Task 6 (component type), Task 7 (damage level), and Task 8 (damage type). Out of the five tasks, three of them (Tasks 1, 5, and 6) are treated as triple classification, while the other two (Tasks 7 and 8) are treated as dual-detection tasks.

Scene level (Task 1) aids in determining the extent of structural damage to an infrastructure. There are three class instances for the scene-level tasks: pixel, object, and structural. Collapse mode (Task 5) helps to ascertain the collapsed mode of an infrastructure, whether it is none-collapse, partial collapse, or global collapse; these form the class instances of Task 5. The component type (Task 6) is associated with identifying the structural part of an infrastructure, which includes beams, columns, walls, and others. Also, damage level (Task 7) provides insight into the extent of damages sustained by infrastructure, and it includes four classes: undamaged, minor, moderate, and heavy. Lastly, damage type (Task 8), which has four classes: undamaged, flexural, shear, and combined, helps to ascertain the type of structural damage associated with an infrastructure. Table 1 gives further details regarding the class instances under the five tasks, and Fig Xa shows samples of the dataset.

By combining information from these inter-tasks (Tasks 1, 5, 6, 7, and 8), civil engineering practitioners and analysts can gain a holistic understanding of the surface structural health of an infrastructure, including the scene level understanding, damage extent, collapse mode, specific structural parts affected, severity of damage, and the nature of structural damage. This multidimensional approach aids in making informed decisions regarding maintenance, repair, or reinforcement measures.

### 4.2 Experimental configuration and implementation

The study experiment is set up in two parts: (1) training and testing with the PEER Hub Image-Net dataset and (2) an application scenario that demonstrates the real-world application aspect of the proposed Guided-DetNet.

Guided-DetNet is implemented using the PyTorch framework on an NVIDIA RTX 3090 with a memory of 32GB. A total of 6,636 samples resized to $640 \times 640$ were used in the experiment. Out of the 6,636 samples, 5,283 samples serve as training samples, and the remaining 1,353 are used as testing samples. Since the proposed GAM in the Guided-DetNet framework generates six additional feature maps given an input, no data augmentation is used to increase the size of the data during the training of Guided-DetNet. However, data augmentation is used to increase the data size from 6,636 to 28,451 for the compared models during training. Nineteen thousand nine hundred sixteen (19,916) augmented samples are used to train the compared models, and 8,535 serve as the testing samples. Guided-DetNet training hyperparameters include a batch size of 32, a learning rate of 0.0001, the Adam optimizer and loss functions, including the Distribution Focal Loss (DFL), CIoU loss, and the binary-cross entropy loss. The DFL and CIoU losses are used for the detection task, and the binary-cross entropy is used for the classification task. In the application scenario test (real-world experiment), a fabricated drone and a drone-based structural health monitoring application [38] are used. The drone-based structural health monitoring application, a graphical user interface (GUI) program, was designed using the PyQt5 library (readers are to refer to our previous studies [33] for additional details regarding the drone-based application). The fabricated drone is connected to the GUI application via a wireless connection, while a human operator flies the drone during the field test. The graphical representation of the configuration and experimental setup is visualized in Fig. 5.

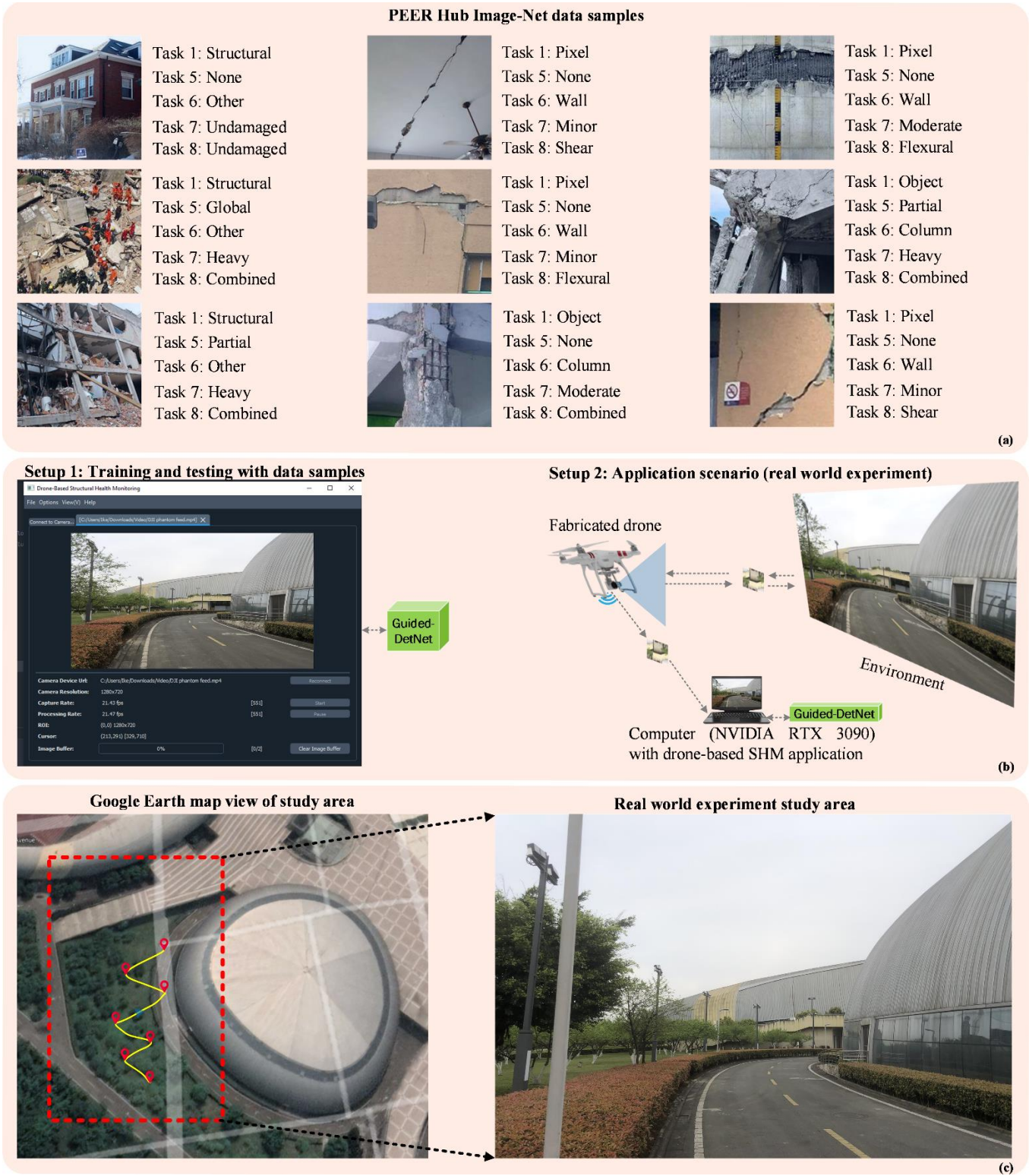

**Fig. 5.** Visual representation of study experimental configuration and setup: (a) samples of the PEER Hub Image-Net dataset with corresponding Task and class instances, (b) the two experimental setups, and (c) Google Earth and scene of the inspection site.

*4.3 Evaluation metrics*

The evaluation metrics used in the study are categorized into two: (1) metrics for the triple classification task and (2) metrics for the dual detection tasks.

In the triple classification task, the training accuracy and losses, precision, recall, F1 Score, normalized confusion matrix, and misclassification rate are used to evaluate Guided-DetNet and the compared classification models. The accuracy and loss will aid in reflecting how well the models adapt to structural damage classification. Precision indicates the models' ability to correctly classify damage without false positives, which is critical in minimizing unnecessary maintenance costs. Recall measures the ability to classify all instances of structural damage, ensuring no critical damage goes undetected. The F1 Score balances precision and recall, emphasizing both accuracy and comprehensiveness. In the dual-detection task inference time, frame per second (fps), normalized confusion matrix, mean average precision (mAP), mean average recall (mAR), and precision-recall curve are used as the evaluation metrics. The inference time and fps evaluate the computational efficiency of the models in real-time applications, which is essential for SHM systems deployed for continuous monitoring. The mAP and mAR assess the accuracy and completeness of detected structural damage; they are calculated using the average precision and average recall based on Intersection over Union (IoU) thresholds (IoU = [0.5, 0.55,…,0.95]), where the IoU computes the offset between the ground truth $gt$ and the predicted bounding boxes $bb$, respectively, as in Eq. (28). The mAP and mAR reflect the model's ability to localize and classify damage under varying conditions, ensuring robustness in detection performance. Also, the precision-recall curve is used to provide a comprehensive view of the trade-off between precision and recall across thresholds, highlighting the models' detection capability under different operational conditions. Further, additional tests are categorized under robustness test and noisy data trained test. In the robustness test, precision is used as the evaluation metric, and accuracy and misdetection rate are used as metrics in noisy data-trained tests; these metrics will aid in reflecting the models' adaptability to real-world scenarios where noise is inevitable. Additionally, under the dual detection task, the visual interpretation of the volumetric contour visual assessment of detected damages is given; this offers a qualitative assessment of the detected damages, aiding practitioners in visualizing the severity and distribution of structural damages for decision-making.

The normalized confusion metrics comprise four parts: (1) True positive (TP), True negative (TN), False positive (FP), and False negative (FN). TP denotes a true classified/detected positive data sample, and TN denotes true classified/detected negative samples. FP implies data samples wrongly classified/detected as positive samples, and FN represents samples wrongly classified/detected as negative data samples.

$$IoU = \frac{area(gt \cap bb)}{area(gt \cup bb)} \quad (28)$$

The other metrics, including precision, recall, accuracy, F1 Score, and misclassification rate, are computed as Eqs. (29) to (33), respectively.

$$precision = \frac{TP}{TP + FP} \quad (29)$$

$$recall = \frac{TP}{TP + FN} \quad (30)$$

$$accuracy = \frac{TP + TN}{TP + TP + FP + FN} \quad (31)$$

$$F1\ score = 2 * \frac{recall * precision}{recall + presicion} \quad (32)$$

$$misclassification\ rate = 1 - accuracy \quad (33)$$

## 5. Results and discussion

Since the proposed Guided-DetNet is capable of classification and detection tasks, the results and discussion section are grouped into two: (1) triple classification and (2) dual detection tasks.

*5.1 Triple classification results and analysis*

The performance of Guided-DetNet in the triple classification task is evaluated and compared against several state-of-the-art models, including EnsembleDetNet [38], YOLOv8-m-cls+PAM [45], EfficientNet-B5+PAM [24], BuildingNet+EEAM+ [28], and ResNeXt-101+CBAM [53]. All the compared models were modified to have a triple classification head for the task. The selected classifiers were embedded with attention models, including CBAM, EEAM+, and PAM, for a fair comparison. The evaluation metrics for the triple classification task include accuracy, precision, recall, F1 score, and misclassification rate.

*5.1.1 Triple classification quantitative results and analysis*

The analysis is based on Fig. 6, which gives a graphical insight into the training accuracy and losses, and Table 2, which provides a quantitative result.

Guided-DetNet showcased satisfactory accuracy, achieving 96%, indicating a high proportion of correctly classified instances. The framework's precision of 1.0 indicates its ability to make positive predictions with high confidence, minimizing the occurrence of false positives. Similarly, Guided-DetNet achieved a commendable recall of 0.92, indicating its capability to identify a substantial portion of relevant instances. The F1 Score of 0.96 signifies an effective balance between precision and recall, highlighting the framework's robustness in handling true positives and false negatives. The low misclassification rate of 0.04 further emphasizes Guided-DetNet's reliability, indicating a minimal rate of misclassified instances; this is a significant advantage in critical applications such as structural damage detection, where misinterpretations can lead to severe consequences. Comparing Guided-DetNet to the selected classifiers, EnsembleDetNet performed well, with an accuracy of 93%, precision of 0.98, and a recall of 0.86; however, Guided-DetNet outperformed it on all metrics, as seen in Table 2 and Fig. 6(c). Guided-DetNet's precision and recall metrics indicate a better balance between identifying relevant instances and avoiding false positives. YOLOv8-m-cls+PAM attained the same precision of 1.0 as Guided-DetNet, indicating both models have a reduced likelihood of false positives, making them more reliable in positive predictions. While YOLOv8-m-cls+PAM achieved a precision of 1.0 and an accuracy of 94%, better than all the state-of-the-art classifiers in this study, Guided-DetNet surpassed it in accuracy by 2%, recall by 0.03, F1 Score, and misclassification rate by 0.02, respectively. Further, a comparison between Guided-DetNet and EfficientNet-B5+PAM, which achieved an accuracy of 91%, precision of 0.88, recall of 0.90, F1 score of 0.89, and a misclassification rate of 0.09 signifies the satisfactory performance of Guided-DetNet. The significant margin in precision and recall compared to EfficientNet-B5+PAM highlights Guided-DetNet's ability to achieve accurate positive predictions while capturing a high proportion of true positives. Lastly, BuildingNet+EEAM+ and ResNeXt-101+CBAM exhibited lower competitive performance compared to Guided-DetNet, particularly in accuracy, precision, and recall. ResNeXt-101+CBAM marginally outperformed BuildingNet+EEAM+ by 1% to 2% on the evaluation metrics.

**Table 2**
Quantitative results for the triple classification task on the PEER Hub Image-Net dataset.

| Model | Accuracy | Precision | Recall | F1 Score | Misclassification rate |
|---|---|---|---|---|---|
| Guided-DetNet | 96 | 1.0 | 0.92 | 0.96 | 0.04 |
| EnsembleDetNet | 93 | 0.98 | 0.86 | 0.92 | 0.07 |
| YOLOv8-m-cls+PAM | 94 | 1.0 | 0.89 | 0.94 | 0.06 |
| EfficientNet-B5+PAM | 91 | 0.88 | 0.90 | 0.89 | 0.09 |
| BuildingNet+EEAM+ | 84 | 0.84 | 0.83 | 0.83 | 0.16 |
| ResNeXt-101+CBAM | 85 | 0.86 | 0.84 | 0.85 | 0.15 |

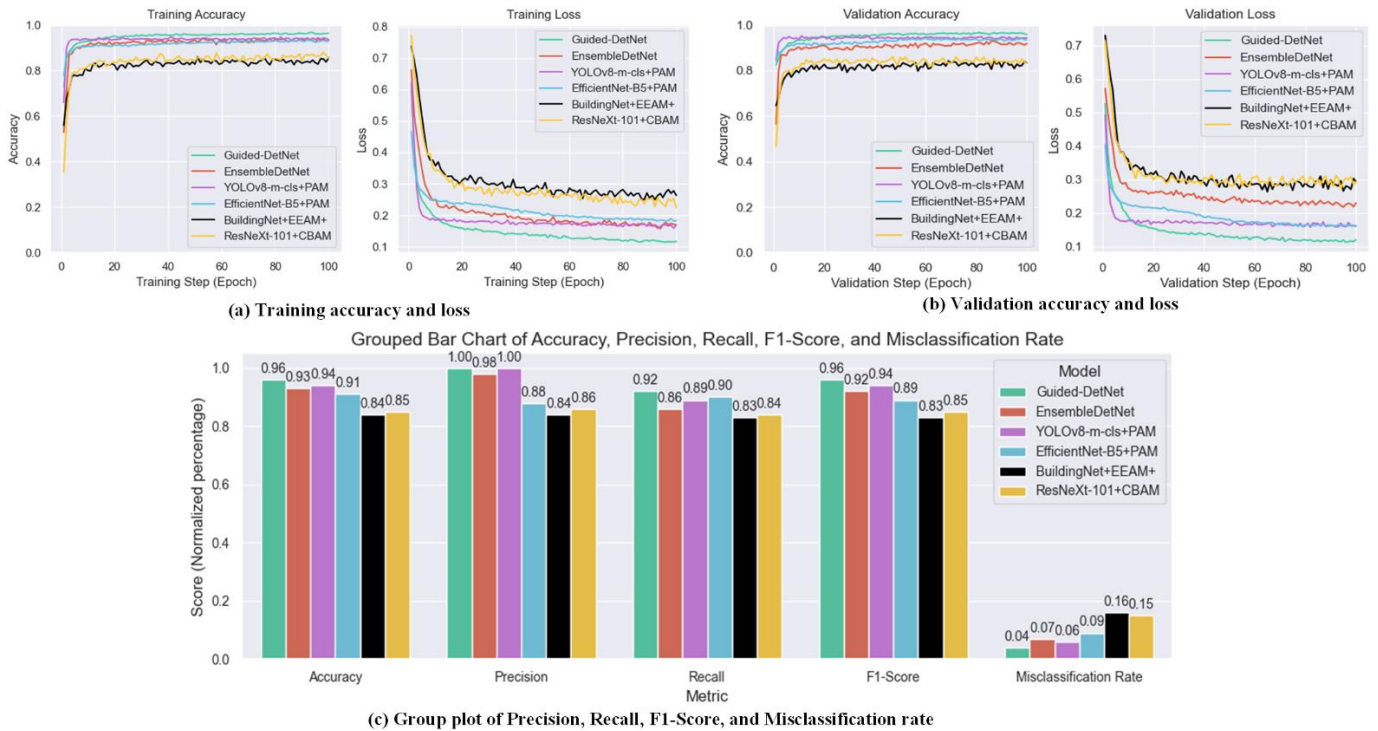

**Fig. 6.** Results of triple classification task: (a) the training accuracies and losses for Guided-DetNet and the compared classifiers, (b) the validation accuracies and loss for Guided-DetNet and the compared classifiers, and (c) the group plot for Guided-DetNet and compared classifiers regarding accuracy, precision, recall, f1-score, and misclassification rate.

The promising performance of Guided-DetNet is influenced by the Generative Attention Module (GAM) and the hierarchical elimination algorithm. The inclusion of GAM in Guided-DetNet significantly contributes to the framework's enhanced performance. GAM introduces multi-varied feature maps encompassing dynamically rotated feature maps and varied fused feature maps that adjust the background and foreground of features, allowing the framework to focus on critical regions relevant to damage detections. This attention mechanism aids in precise localization and feature saliency, which is crucial for accurate predictions. Also, the hierarchical elimination algorithm leverages the hierarchical relationship between tasks, allowing the framework to refine class instances effectively. The hierarchical elimination algorithm, as outlined in Algorithm 1, plays a pivotal role in reducing noisy labels and refining class instances. By considering the relationships between tasks (e.g., scene level, collapse mode, component type), the algorithm dynamically adjusts class instances based on the predictions of preceding tasks. This hierarchical approach ensures that the subsequent tasks benefit from more accurate and coherent predictions, contributing to the overall success of Guided-DetNet.

*5.1.2 Normalized confusion matrix for triple classification*

The normalized confusion matrix provides valuable insights into the performance of the models in the triple classification task, where the classes are associated with scene level (Task 1), collapse mode (Task 5), and component type (Task 6). The analysis considers the class instances for each task: Task 1 (scene level) with classes [Pixel, Object, Structural], Task 5 (collapse mode) with classes [None, Partial, Global], and Task 6 (component type) with classes [Beam, Column, Wall, Other].

Guided-DetNet exhibits high accuracy in Task 1 in distinguishing pixel-level and object-level views, as evidenced by a 100% prediction attained for these classes, as seen in Fig. 7(a). Similarly, Guided-DetNet excels in identifying structural-level scenes by predicting 94% correct. In Task 5, Guided-DetNet demonstrates good performance in correctly identifying instances with no collapse at 100%. For partial and global instances, Guided-DetNet correctly predicted 95% and 96%, respectively. In Task 6, a moderately challenging task, Guided-DetNet correctly predicted 91% for the beam and the other class instance and 95% for the wall instance. The lowest correct prediction score by Guided-DetNet from Fig. 7a is 89% for the column class instance. The inclusion of the Generative Attention Module in Guided-DetNet has a significant impact on the framework's ability to focus on important features, contributing to the satisfactory performance of the classification of the various class instances. This is particularly evident in the precise classification of scene level and good classification of collapse modes and component types. Also, the hierarchical elimination algorithm ensures that the framework leverages information from Task 1 to guide the classification in subsequent tasks; this contributes to the coherent classification of collapse mode and component type class instances, respectively.

A comparative analysis of the other classifiers provides insights into their strengths and weaknesses in handling the triple classification task. From Fig. 7, EnsembleDetNet exhibits good performance overall; EnsembleDetNet exhibits slightly fewer incorrect predictions on some class instances of Tasks 5 and 6. YOLOv8-m-cls+PAM shows satisfactory performance, but the attention-enhancing capabilities and hierarchical elimination algorithm of Guided-DetNet lead to more precise and correct predictions. EfficientNet-B5+PAM demonstrates slightly lower accuracy on some class instances in Tasks 5 and 6, particularly non-collapse and beam class instances. BuildingNet+EEAM+ exhibits a moderately lower accuracy in most of Task 6 class instances, as seen in Fig. 7e, indicating a challenge in classifying Task 6 instances. Similarly, ResNeXt-101+CBAM also recorded a moderately lower classification accuracy on three out of four class instances of Task 6. Overall, the comparative analysis of the compared classifiers from Fig. 7 shows that the lowest correct classification score for EnsembleDetNet is 75%, YOLOv8-m-cls+PAM is 80%, EfficientNet-B5+PAM is 77%, BuildingNet+EEAM+ is 69%, and ResNeXt-101+CBAM is 67% compared to Guided-DetNets lowest classification accuracy of 89%.

The normalized confusion matrix analysis highlights Guided-DetNet's promising performance in the triple classification task, emphasizing the positive impact of the Generative Attention Module and hierarchical elimination algorithm. The framework excels in the precise classification of scene levels, collapse modes, and structural component types, showcasing its potential for robust and accurate multi-task structural damage classification in diverse scenarios.

*5.2 Dual detection results and analysis*

Guided-DetNet is compared with EnsembleDetNet, YOLOv8-m-det+PAM [45], EfficientDet+PAM [30], YOLOX-L+CBAM [50], YOLOv5-m+CBAM [47], YOLOv7-X+CBAM [54], CSPFaster-R-CNN+EEAM+ [55], and ResNet-34-VFNet+EEAM+ in the dual detection task. The evaluation metrics used include inference time, frame processing speed, normalized confusion matrix, AP, mAP, mAR, precision-recall, and F1 Score. Additionally, robustness and the volumetric assessment of detected damages are discussed.

*5.2.1 Dual detection quantitative results and analysis*

From the quantitative results tabulated in Table 3 and the graphical results of Fig. 8, Guided-DetNet attained promising results across various metrics, showcasing its competence in dual detection tasks. The framework attains a precision of 0.94, indicating a low rate of false positives, which is crucial in structural damage detection where accuracy is essential. A recall of 0.86 demonstrates the framework's ability to identify a substantial portion of actual positive instances. The F1 Score of 0.90 signifies a balanced performance between precision and recall, highlighting Guided-DetNet's effectiveness in handling true positives and false negatives. Guided-DetNet achieved an average precision (AP) of 0.91, emphasizing its capability to provide accurate localization and classification of class instances. The mean average precision (mAP) of 0.79 further underscores the framework's robustness in handling multiple classes simultaneously. The mean average recall (mAR) of 0.76 indicates a high overall recall performance, showcasing Guided-DetNet's ability to detect objects comprehensively. In terms of the number of frames the framework can process and the time taken to process a single input, Guided-DetNet attained an average frame per second (fps) of 57.10s and an average inference time of 6.17m/s (milliseconds) from the three repeated fps and inference test, as seen from Fig. 9. With an average fps of 57.10s, Guided-DetNet demonstrates its applicability in real-time structural health monitoring applications. Additionally, an average inference of 6.17 m/s solidifies the framework's suitability for applications requiring quick inferencing. The satisfactory and promising performance of Guided-DetNet is attributed to the proposed modules in the framework. Guided-DetNet's integration with GAM significantly contributes to its promising performance. GAM enhances and increases feature diversity, improving precise localization and classification, which is vital for detection tasks. Also, the hierarchical elimination algorithm plays a crucial role in refining class instances and reducing noisy labels. The algorithm leverages the relationships between tasks (Tasks 5, 7, and 8), contributing to more accurate predictions in dual detection tasks. Fig. 10 shows samples of detection results on the testing dataset and the real-world experiment (field test).

Guided-DetNet distinguishes itself from the compared state-of-the-art detectors, exhibiting superior performance across multiple metrics. Notably, against EnsembleDetNet, Guided-DetNet excels in precision, recall, F1 Score, AP, mAP, and mAR, mostly with an average difference of 4%. Also, when compared with YOLOv8-m-det+PAM, the most promising detector among the compared detectors, Guided-DetNet emerges as the more promising detector, showcasing higher precision, AP, mAP, recall, and F1 Score. Guided-DetNet attained approximately 3% higher performance on the majority of the metrics compared with YOLOv8-m-det+PAM. Further, compared with EfficientDet-D5+PAM, Guided-DetNet maintains its superiority in precision, recall, F1 Score, AP, and mAP, mostly with a difference of more than 6% on most of the metrics. The comparative analysis extends to state-of-the-art YOLO-based detectors, including YOLOX-L+CBAM, YOLOv5-m+CBAM, and YOLOv7-X+CBAM, among the YOLO-based detectors, YOLOv7-X+CBAM attained the most satisfactory results as seen from Table 3 and Figs. 8 and 9 on varying metrics. Guided-DetNet attained a difference of 2% to 5% more in the metrics, as seen in Table 3, compared to YOLOv7-X+CBAM.

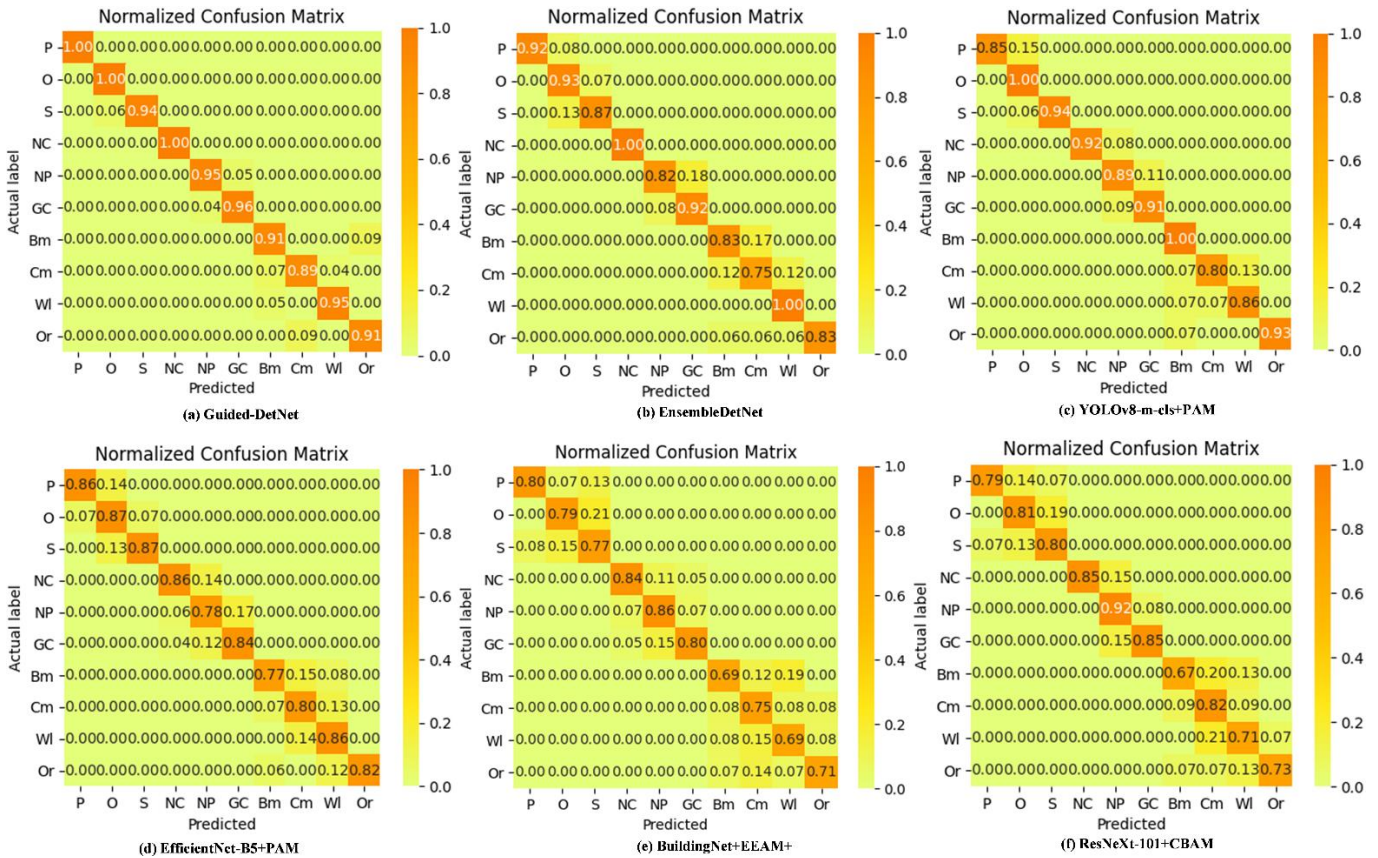

**Fig. 7.** Comparison of a normalized confusion matrix for Guided-DetNet and the compared classifiers for the triple classification task.

**Table 3**
Quantitative results of Guided-DetNet and the compared detectors for the dual detection task on the PEER Hub Image-Net dataset.

| Model | Precision | Recall | F1 Score | AP | mAP | mAR | FPS | Inference time |
|---|---|---|---|---|---|---|---|---|
| Guided-DetNet | 0.94 | 0.86 | 0.90 | 0.91 | 0.79 | 0.76 | 57.04 | 6.22 |
| EnsembleDetNet | 0.90 | 0.83 | 0.86 | 0.85 | 0.75 | 0.74 | 63.42 | 16.04 |
| YOLOv8-m-det+PAM | 0.91 | 0.83 | 0.87 | 0.86 | 0.76 | 0.74 | 72.13 | 2.08 |
| EfficientDet-D5+PAM | 0.80 | 0.79 | 0.79 | 0.78 | 0.70 | 0.68 | 33.05 | 17.34 |
| YOLOX-L+CBAM | 0.86 | 0.82 | 0.84 | 0.81 | 0.72 | 0.70 | 70.21 | 11.33 |
| YOLOv5-m+CBAM | 0.78 | 0.77 | 0.77 | 0.74 | 0.68 | 0.66 | 119.03 | 4.30 |
| YOLOv7-X+CBAM | 0.90 | 0.85 | 0.87 | 0.85 | 0.74 | 0.74 | 109.04 | 7.22 |
| CSPFaster-R-CNN+EEAM+ | 0.74 | 0.66 | 0.70 | 0.71 | 0.61 | 0.59 | 9.12 | 115.56 |
| ResNet-34-VFNet+EEAM | 0.70 | 0.63 | 0.66 | 0.68 | 0.58 | 0.55 | 11.31 | 107.44 |

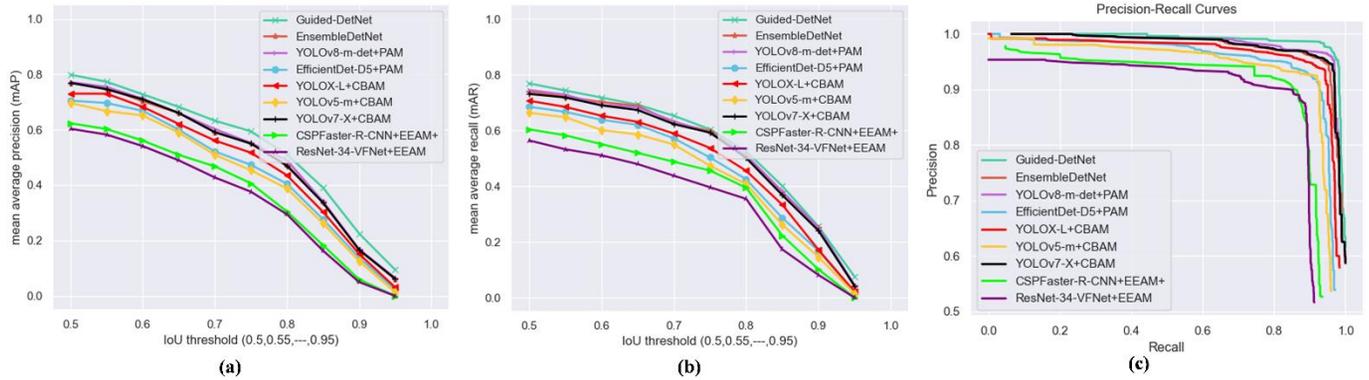

**Fig. 8.** Dual detection results for Guided-DetNet and the compared detectors: (a) mean average precision, (b) mean average recall, and (c) precision-recall curve.

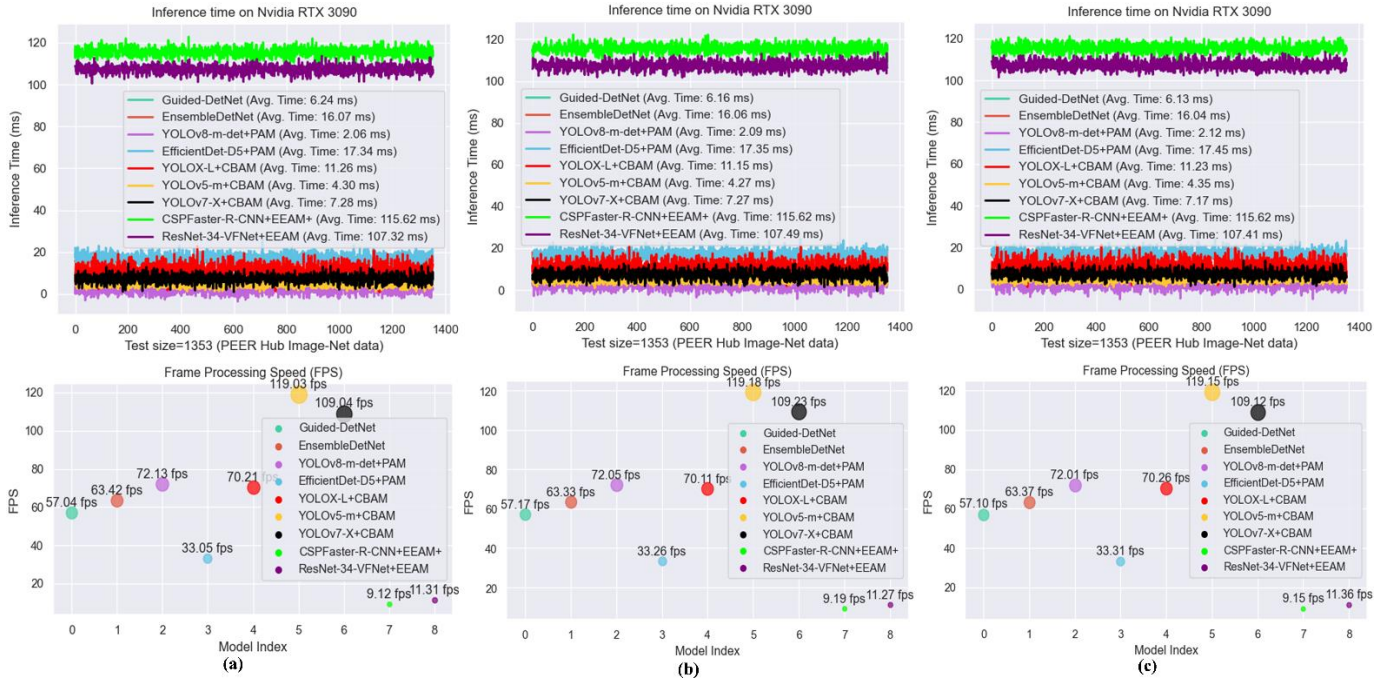

**Fig. 9.** Guided-DetNet and the compared detectors inference time and frame per second test results; this was repeated three times to observe consistency in results.

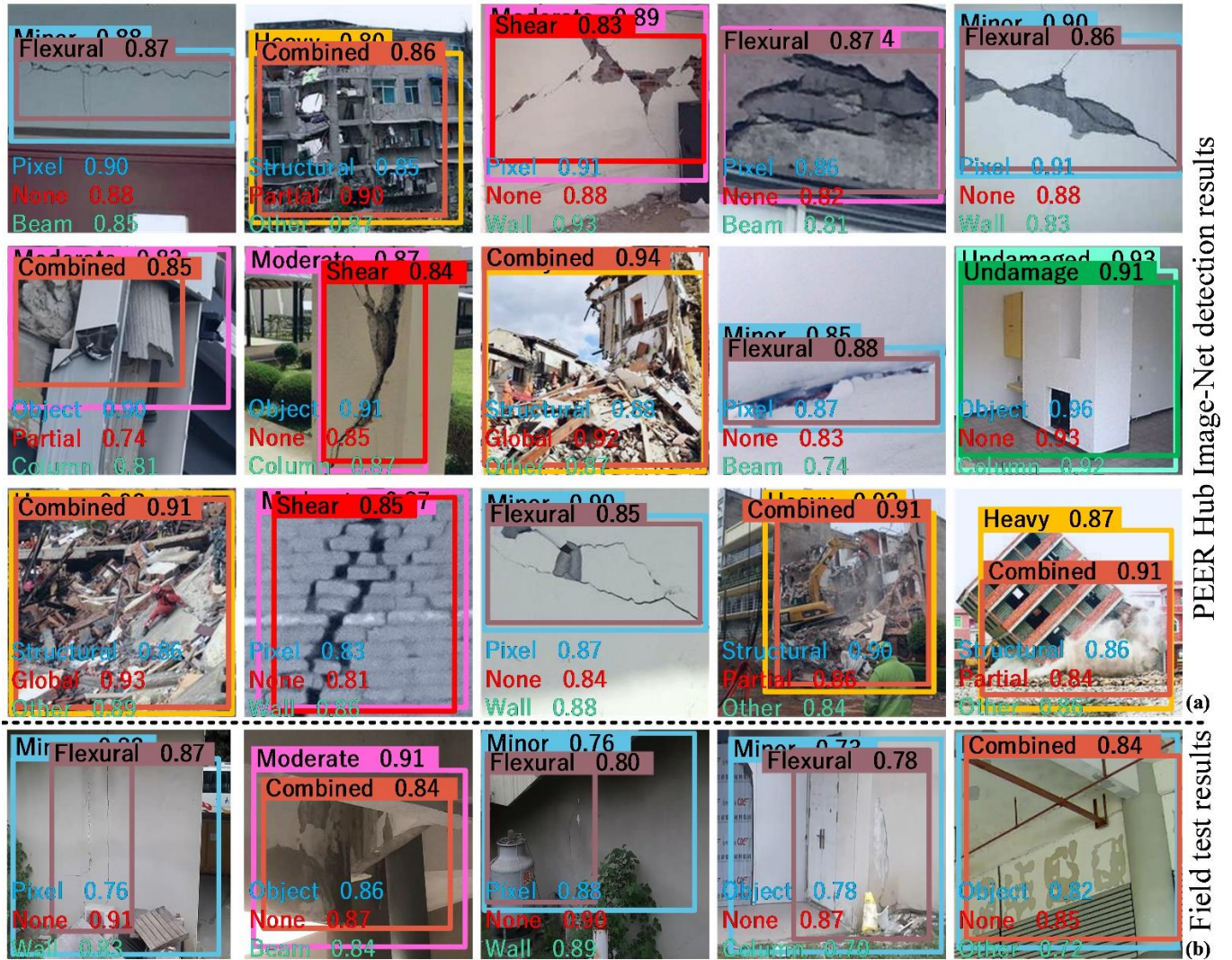

**Fig. 10.** Dual detection and triple classification results for Guided-DetNet: (a) results on PEER Hub Image-Net test data samples and (b) field test results.

For the least performed detectors, CSPFaster-R-CNN+EEAM+ and ResNet-34- VFNet+EEAM, Guided-DetNet maintains its dominance, highlighting its effectiveness in dual detection. Guided-DetNet attains an average difference of more than 17% on most metrics compared to the two least-performed detectors. The consistently superior performance across diverse metrics underscores Guided-DetNet's versatility and potential for applications demanding precise and efficient object detection and localization. Although Guided-DetNet attained promising performance compared to all the detectors on most metrics, in terms of inference time and fps, as seen from Table 3 and Fig. 9, the YOLO-based models are superior regarding the fps and the inference time. Among the YOLO-based detectors, YOLOv5-m+CBAM attained the best fps and inference time.

In conclusion, Guided-DetNet emerges as a formidable framework in dual detection tasks, outperforming state-of-the-art models in terms of precision, recall, F1 Score, AP, mAP, and mAR. Also, Guided-DetNet attained competitive results regarding fps and inference time. The incorporation of GAM and the hierarchical elimination algorithm contributes to the framework's effectiveness in handling dual detection tasks. Guided-DetNet demonstrates its potential for real-world applications where accurate and efficient object detection and localization are critical.

*5.2.2 Normalized confusion matrix analysis on dual detection*

A normalized confusion matrix is used to assess and pinpoint the precision of Guided-DetNet and the compared detectors with a focus on the class instances of the dual detection task (Tasks 7 and 8).

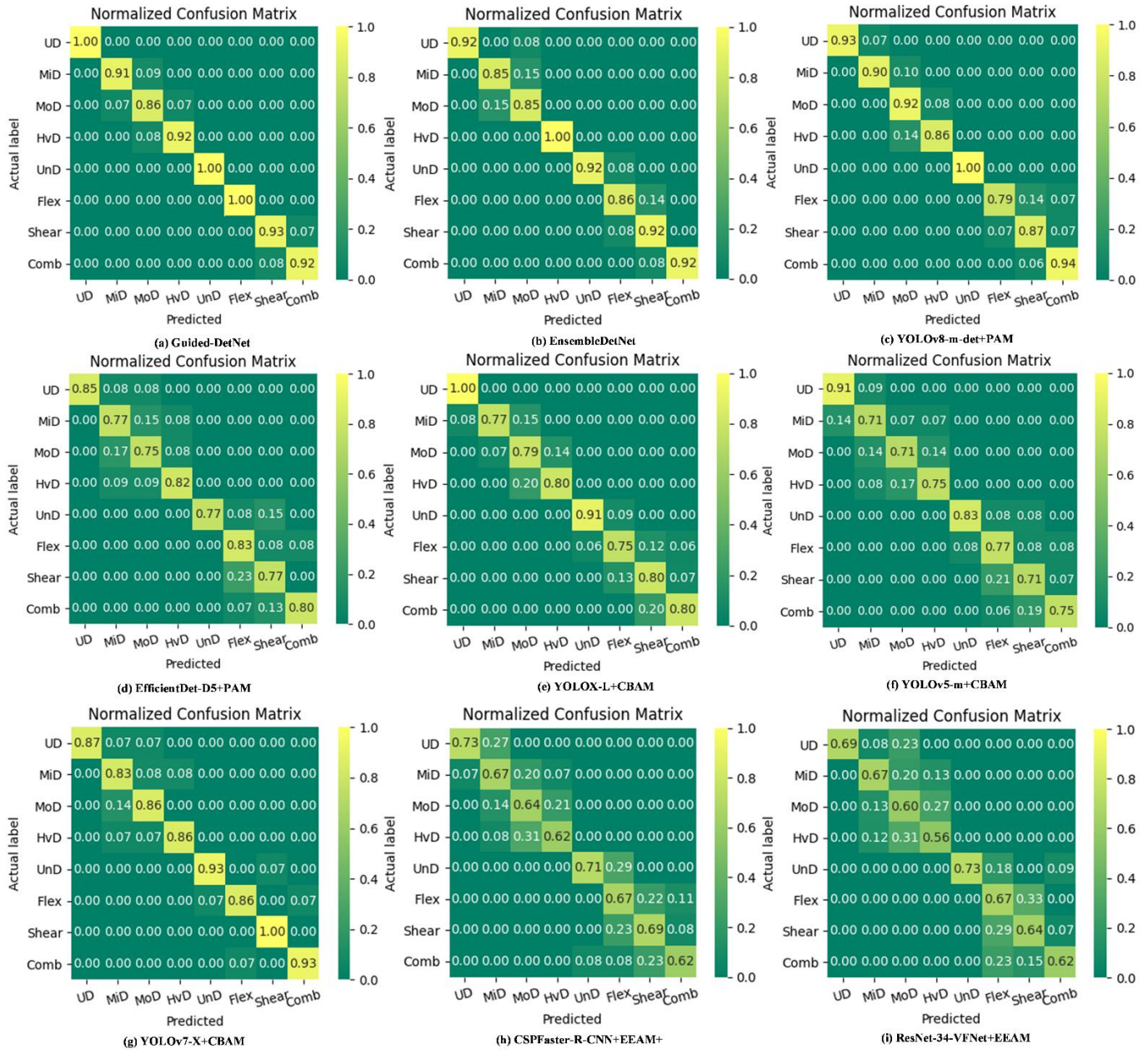

**Fig. 11.** Dual detection normalized confusion results for Guided-DetNet and the compared detectors.

From Fig. 11a, Guided-DetNet showcases notable precision and recall across categories, including undamaged, minor, moderate, and heavy for Task 7 (damage level). This indicates a balanced performance in accurately categorizing the extent of damage, ranging from minor defects to severe structural damages. Task 8 (damage type) further shows Guided-DetNet's efficacy, particularly in distinguishing between damage types. From Fig. 11a, the lowest correct predicted instance of Guided-DetNet is 86%, which is for the moderate class instance in Task 7, and the highest predicted score is 100% for classes undamaged in Task 7 and undamaged and flexural in Task 8. The Guided-DetNet prediction for the other class instances in Tasks 7 and 8 falls within the 90[th] percentile, as seen in Fig. 11a. The diagonal scores of the confusion matrix (Fig. 11a), representing true positive rates for each class, demonstrate Guided-DetNet's ability to detect instances of interest accurately. Moreover, the non-diagonal scores shed light on potential mis-detection, and a closer inspection reveals that the mis-detection is generally balanced across classes, indicating a well-balanced trade-off between false positives and false negatives.

Examining the normalized confusion matrix results from Fig. 11 for the compared detectors shows EnsembleDetNet demonstrates competitive performance, achieving commendable precision and recall for undamaged and heavy damage class instances in Task 7. However, its performance reduces for minor and moderate instances, indicating challenges with the two class instances. EnsembleDetNet displayed proficiency in identifying undamaged, shear, and combined damage instances but encountered minor difficulties in distinguishing flexural damage. Among the YOLO-based detectors, YOLOv8-m-det+PAM showcases more robust performance than the other three YOLO-based detectors. YOLOv8-m-det+PAM predicted undamaged, minor, and moderated instances in Task 7 with high precision and recall. However, the detector experienced a reduction in accurately predicting heavy instances, reflecting limitations in discerning more severe structural damages. In Task 8, YOLOv8-m-det+PAM excelled in identifying undamaged and combined damage instances but experienced a decrease in precision and recall for flexural and shear. The other three YOLO-based models attained similar but varying performances; in terms of performance ranking, YOLOv7-X+CBAM precedes YOLOX-L+CBAM and YOLOX-L+CBAM precedes YOLOv5-m+CBAM. EfficientDet-D5+PAM demonstrates promising precision and recall. However, its performance mostly falls between the 70$^{th}$ and 80$^{th}$ percentile on all the class instances for Tasks 7 and 8, arguably highlighting challenges in accurately categorizing Tasks 7 and 8 class instances with more precision. From Fig. 11, the two least performed detectors, CSPFaster-R-CNN+EEAM+ and ResNet-34-VFNet+EEAM, exhibit challenges in precision and recall across various class instances in both Task 7 and Task 8. CSPFaster-R-CNN+EEAM+ struggles particularly in accurately predicting minor, moderate, and heavy instances in Task 7, indicating limitations in identifying structural damages. In Task 8, CSPFaster-R-CNN+EEAM+ encounters challenges in distinguishing between damage types, resulting in lower precision and recall for flexural, shear, and combined damage. ResNet-34-VFNet+EEAM faces similar limitations, demonstrating lower precision and recall for all class instances in Task 7. The detector's performance diminished significantly for heavy class instances. In Task 8, ResNet-34-VFNet+EEAM struggles to distinguish between damage types, resulting in lower precision and recall for shear, flexural, and combined damage. The model's overall performance in the dual detection task falls behind that of more advanced detectors.

The comparative analysis indicates Guided-DetNet's superiority in precision and recall across all class instances in Tasks 7 and 8 compared to other detectors. Guided-DetNet consistently outperformed its counterparts, showcasing its capability to categorize damage levels and types accurately. The highest predicted scores for various class instances further emphasize Guided-DetNet's understanding of structural damages, positioning it as a robust solution for applications requiring reliable detections. Overall, the detailed analysis reaffirms Guided-DetNet's promising performance in structural damage detection.

*5.2.3 Results and analysis of noisy labeled data*

Guided-DetNet is further tested for robustness by training data with noisy labels. Out of the initial 5,283 training samples, 1,000 samples were randomly selected and noisily labeled; afterward added back to the remaining samples, summing up to 5,283; this is then used to train Guided-DetNet and the other detectors. The experiment aimed to assess Guided-DetNet and the other detectors in scenarios where the training data included noisy labels.

Guided-DetNet emerged as the best-performing detector in the noisy labeled data test, as seen in Figs. 12 and 13, achieving an accuracy of 95%, mAP of 0.77, and mAR of 0.74. Guided-DetNet showcases its ability to navigate through the complexities introduced by noisy labels. The framework's minimal misdetection rate of 0.05 indicates its efficacy in maintaining a low false-positive rate despite the challenges posed by inaccurate annotations. The hierarchical elimination algorithm plays a crucial role in this context by refining class instances based on the intertask relationships and ensuring coherent predictions even in the presence of labeling inaccuracies. GAM further contributes to the adaptability of Guided-DetNet via the induced diversified features, enabling the framework to focus on important features.

Comparatively, the other detectors also demonstrate resilience to noisy labels but with varying degrees of success. EnsembleDetNet, YOLOv8-m-det+PAM, and YOLOv7-X+CBAM attained accuracies of 90%, 91%, and 87%, respectively, compared to Guided-DetNet's 95%. While these detectors exhibit commendable performance, their misdetection rate falls between 9% and 13%, as evident in Fig. 13. In terms of mAP and mAR, EnsembleDetNet, YOLOv8-m-det+PAM, and YOLOv7-X+CBAM attained between 72% and 74%. Comparing the mAP and mAR of EnsembleDetNet, YOLOv8-m-det+PAM, and YOLOv7-X+CBAM to Guided-DetNet indicates Guided-DetNet's superior ability to maintain a high detection rate, indicating its robustness and adaptability in handling noisy datasets. In the middle tier, YOLOX-L+CBAM, EfficientDet-D5+PAM, and YOLOv5-m+CBAM demonstrate moderate performance, with accuracies ranging from 76% to 86% and misdetection ranging from 14% to 24%. For mAP and mAR, YOLOX-L+CBAM, EfficientDet-D5+PAM, and YOLOv5-m+CBAM attained between 64% and 70%. These detectors exhibit a reasonable ability to cope with noisy labels, but the results suggest a moderate performance compared to Guided-DetNet. On the contrary, CSPFaster-R-CNN+EEAM+ and ResNet-34-VFNet+EEAM faced challenges in handling noisy data, recording lower accuracies of 69% and 66%, respectively. The higher misdetection rates of 0.31 and 0.34 indicate a notable struggle to maintain a low false-positive rate in the presence of labeling inaccuracies. Also, these two least-performing detectors recorded the least mAP and mAR, which fall between 50% and 57%.

The results from Figs. 12 and 13 reaffirm Guided-DetNet as a robust and reliable detector for structural damage detection in scenarios where datasets contain noisy labels. The results show the importance of incorporating attention mechanisms and task-related information in enhancing a detector's ability to handle the complexities introduced by noisy labeled data, further emphasizing Guided-DetNet's effectiveness in challenging and dynamic environments.

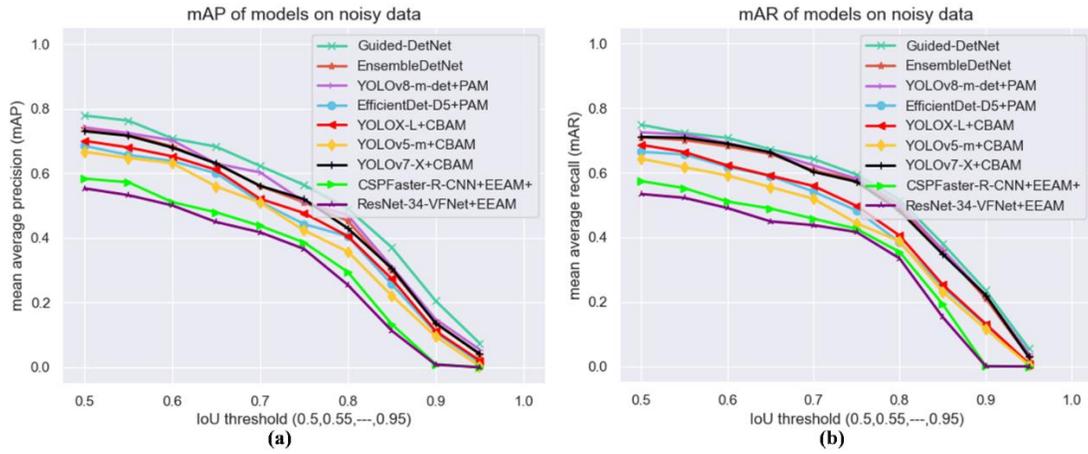

**Fig. 12.** Mean average precision (mAP) and mean average recall (mAR) of Guided-DetNet and the compared detectors on noisy labeled PEER Hub Image-Net dataset.

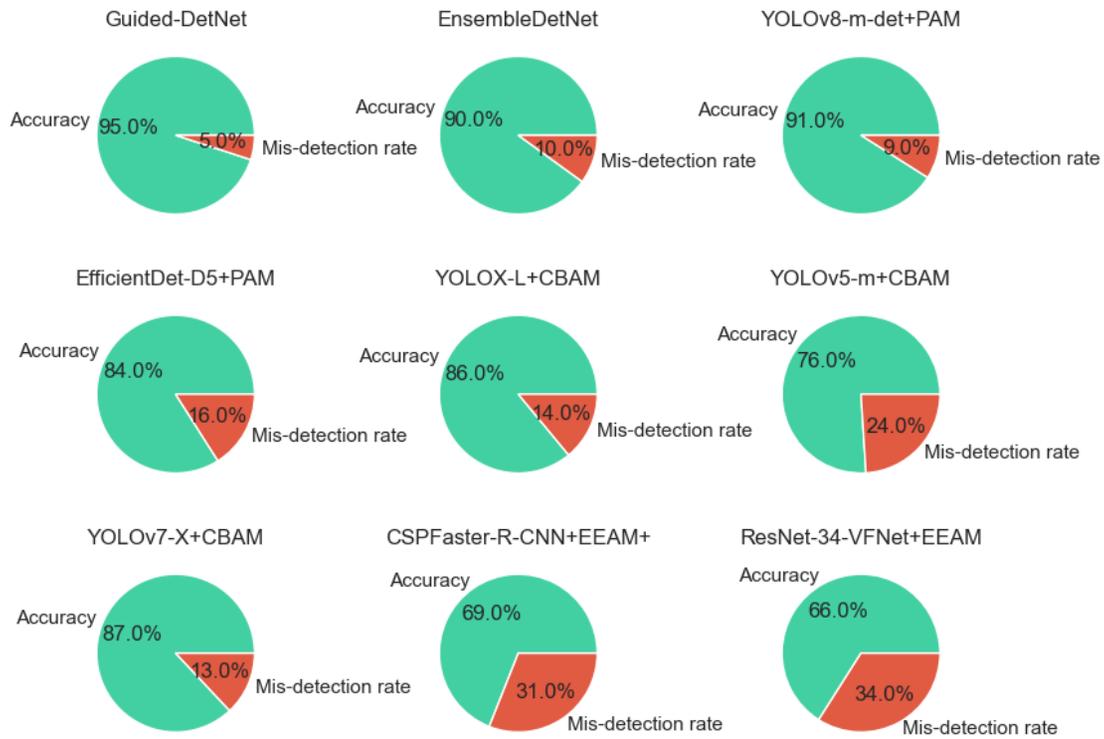

**Fig. 13.** Accuracy and mis-detection rate of Guided-DetNet and the compared detectors on noisy labeled PEER Hub Image-Net dataset.

*5.2.4 Robustness test and analysis under varying conditions*

The robustness of structural damage detection models is crucial for their real-world applicability. This section analyzes the performance of Guided-DetNet and the compared detectors under varying conditions, specifically: (1) background variation, (2) occlusion, and (3) rainy-snowy conditions.

Under the background variation test, which fuses two images based on intensity levels set at 0.2, 0.4, and 0.6, corresponding to low, moderate, and high intensities, respectively, Guided-DetNet demonstrates robust performance. Guided-DetNet achieved a precision score of 0.92 for low intensity. As the background variation intensity increases to a moderate level (0.4), Guided-DetNet maintains its superiority with a precision score of 0.88. Under high-level (0.6) background variation, Guided-DetNet remains resilient, recording a precision score of 0.83; this emphasizes the framework's capacity to handle more complex and diverse background scenarios, ensuring accurate structural damage detection. Other highly-performed competitive models, including EnsembleDetNet, YOLOv8-m-det+PAM, and YOLOv7-L+CBAM, exhibited commendable performances, showcasing their ability to discern structural damage amidst varied backgrounds. From Fig. 14, EnsembleDetNet, YOLOv8-m-det+PAM, and YOLOv7-L+CBAM continue to deliver competitive performances but with a slight decline in precision in moderate and high intensities, respectively. YOLOX-L+CBAM, EfficientDet-D5+PAM, and YOLOv5-m+CBAM continue to perform as moderate competitive detectors under varying intensity levels, while CSPFaster-R-CNN+EEAM+ and ResNet-34-VFNet+EEAM recorded the least precisions.

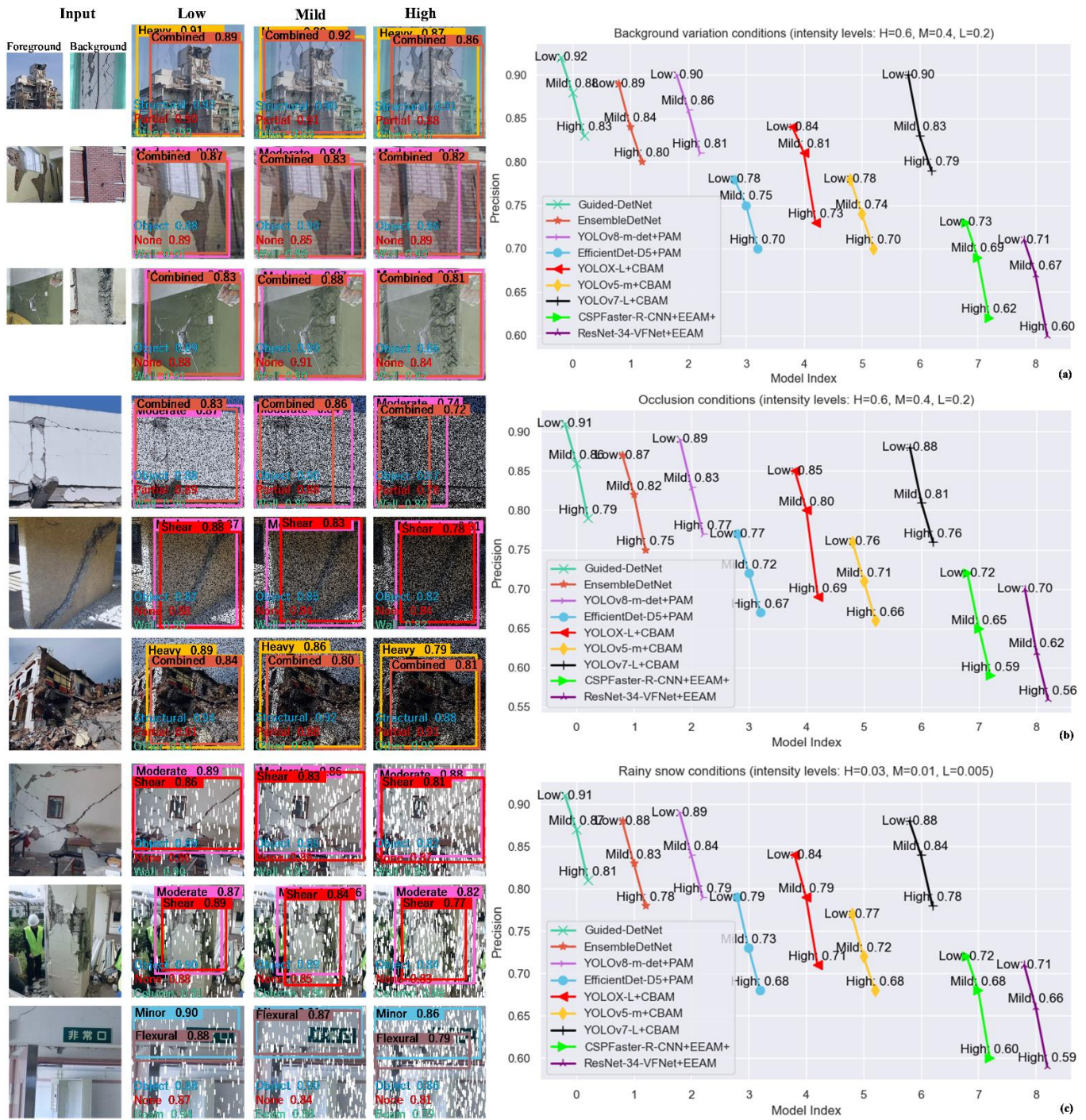

**Fig. 14.** Guided-DetNet and the compared detectors robustness test under varying environmental conditions: (a) background variation, (b) occlusion condition, and (c) rainy-snowy conditions.

In the occlusion condition, intensities are set at 0.2, 0.4, and 0.6 for low, moderate, and high, respectively. At low occlusion intensity, Guided-DetNet excels with a precision score of 0.91, showcasing its capability to identify and effectively categorize structural damage instances even when partially obscured. As occlusion intensity increases to a moderate level, Guided-DetNet maintains a high precision score of 0.86, suggesting the framework's robustness in dealing with moderately obscured structural elements and contributing to reliable damage detection. Under high occlusion intensity, Guided-DetNet continues to outperform other models with a precision score of 0.79; this highlights Guided-DetNet's effectiveness in detecting structural damage instances even when significant portions are occluded. EnsembleDetNet, YOLOv8-m-det+PAM, and YOLOv7-L+CBAM, the highly performed competitive detectors, consistently exhibit commendable precision scores across all occlusion intensity levels. At low intensity, these models maintain precision scores of 0.87, 0.89, and 0.88, respectively, showcasing their robustness in accurately detecting structural damage under partial occlusion. As occlusion intensity increases, these models exhibit a moderate decline in precision but remain competitive. YOLOv8-m-det+PAM stands out with a precision score of 0.77 under high occlusion intensity, emphasizing its resilience in challenging scenarios. YOLOX-L+CBAM, EfficientDet-D5+PAM, and YOLOv5-m+CBAM demonstrate moderate precision scores across occlusion intensity levels. These detectors exhibit precision scores ranging from 0.76 to 0.85 at low intensity and demonstrate a gradual decline with increasing occlusion. The performance of moderately performed detectors suggests their suitability for scenarios with moderate occlusion, where accurate damage detection remains feasible. On the contrary, CSPFaster-R-CNN+EEAM+ and ResNet-34-VFNet+EEAM continued to record the least precisions between 0.56 and 0.72 across all intensity levels, as evident in Fig. 14.

In rainy-snowy conditions, where intensities are set as low (0.005), moderate (0.01), and high (0.03), Guided-DetNet and the compared detectors demonstrated similar performances but varying precisions as compared to the background variation and occlusion condition, respectively. Under low rainy-snowy intensity, Guided-DetNet attained the highest precision score of 0.91, and the highly performed competitive detectors, including EnsembleDetNet, YOLOv8-m-det+PAM, and YOLOv7-L+CBAM, attained commendable precision scores of 0.88, 0.89, and 0.88, respectively. Moderately performed detectors, namely YOLOX-L+CBAM, EfficientDet-D5+PAM, and YOLOv5-m+CBAM, maintain moderate precision scores, ranging from 0.76 to 0.85, indicating adaptability to mild weather challenges. However, the detectors that performed the least, CSPFaster-R-CNN+EEAM+ and ResNet-34-VFNet+EEAM, recorded lower precisions of 72 and 71, respectively. A gradual declining performance is observed in Guided-DetNet and the highly performed detectors when intensity is increased from moderate level to high level.

Overall, Guided-DetNet proves to be a robust and reliable detector for structural damage detection in challenging environmental conditions, with promising implications for practical applications.

### 5.2.5 Volumetric contour visual assessment post-detection

The utilization of Volumetric Contour Visual Assessment (VCVA) within Guided-DetNet significantly enhances the analysis of detected damages by providing quantification of the severity of detected damages. The integration of VCVA allows for a comprehensive examination of the spatial distribution and extent of damages in three dimensions (3D), contributing valuable insights for post-detection analysis.

By leveraging VCVA, Guided-DetNet extends its capabilities beyond mere detection, offering additional means to assess the severity and impact of detected damages within critical infrastructure, as seen in Fig. 15. The visual representation of damages through VCVA aims to provide civil engineering practitioners with a more detailed understanding of the affected areas, facilitating informed decision-making in the context of maintenance and management. The detailed visualization enabled by VCVA facilitates a more precise understanding of the damages' spatial characteristics, allowing civil engineering practitioners to identify critical areas that require immediate attention. With VCVA providing a severity (intensity) level, a score, and a 3D representation of the damages, as seen in Fig. 15, VCVA can be categorized as an advanced level of post-detection analysis, which is particularly valuable in critical infrastructure scenarios where the accurate evaluation of damages can significantly influence decision-making processes. Moreover, the quantification of severity using VCVA potentially aligns with industry best practices, enabling a standardized and objective assessment of damages.

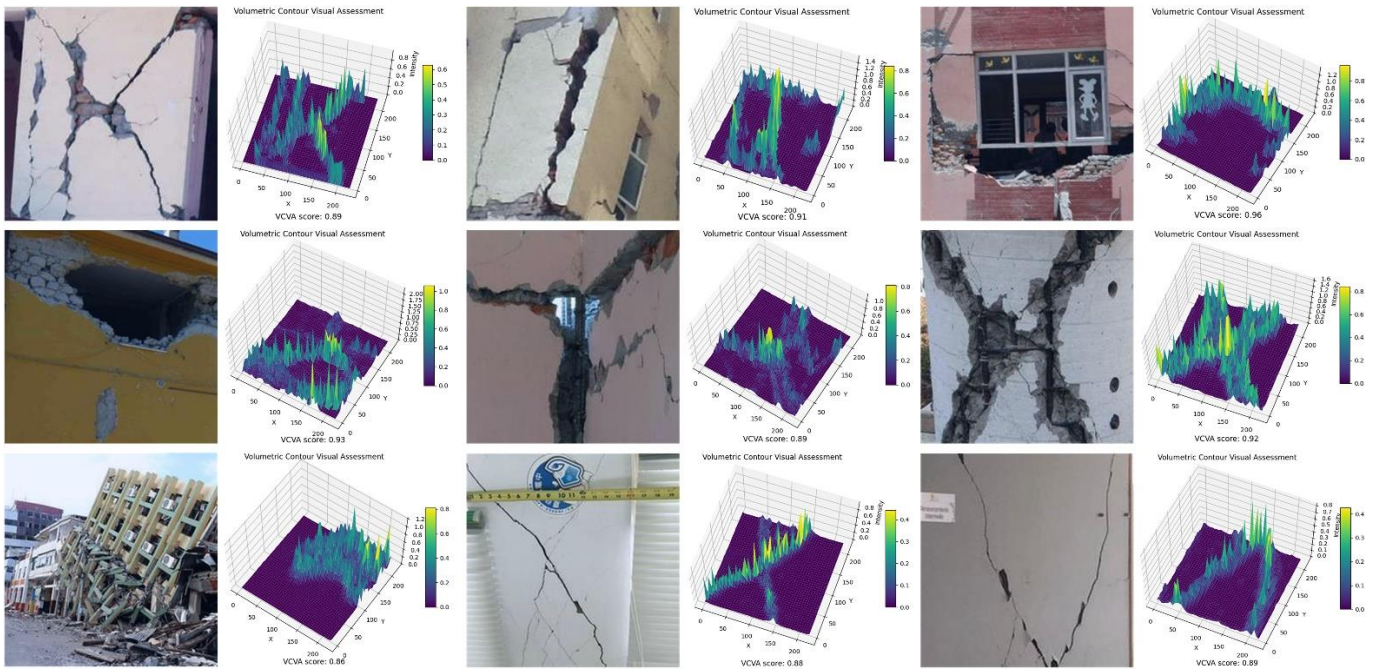

**Fig. 15.** Guided-DetNet 3D representation and volumetric quantification of detected damages.

In conclusion, the incorporation of VCVA within Guided-DetNet elevates the analysis of detected damages by providing an advanced 3D-focused quantification of severity. This advanced post-detection assessment contributes to more informed decision-making in the maintenance and management of critical infrastructure, ensuring that resources are strategically allocated based on a refined understanding of the damages' spatial distribution and impact.

*5.3 Real-world application of Guided-DetNet*

The real-world application of Guided-DetNet and the compared detectors involves the use of a fabricated drone, which establishes a wireless connection with a GUI application, as illustrated in Fig. 5b. The experiment is conducted in two phases: (1) distance to detection and (2) detection under drone speed. In the distance-to-detection setup, the experimental setup considers three distance scenarios – 15 meters, 30 meters, and 50 meters – to evaluate the detectors' effectiveness at different proximities and scales of varying damages. In the detection under drone speed setup, the drone executes a zig-zag flight pattern (refer to Fig. 5b) at different speeds: normal speed: 10 km/h, moderate speed: 20 km/h, and high speed: 30 km/h.

*5.3.1 Distance to detection analysis*

Evaluating the performance of structural damage detection models at different distances is crucial for understanding their effectiveness across varying proximity scenarios as well as detecting objects of varying scales. The precision scores obtained under short (15 meters), medium (30 meters), and long (50 meters) distances give insight into each detector's ability to identify damages from varying distances.

From Fig. 16a, Guided-DetNet consistently exhibits promising performance across all distance tests. With precision scores ranging from 0.87 to 0.94, Guided-DetNet demonstrates satisfactory accuracy in detecting structural damages, even from long distances. This high precision suggests Guided-DetNet's robustness in capturing fine details, making it well-suited for applications that require long-distance drone inspections. EnsembleDetNet maintains competitive precision scores across short, medium, and long distances, with scores ranging from 0.80 to 0.91. YOLOv8-m-det+PAM demonstrates commendable precision across distance tests, with scores ranging from 0.77 to 0.91. Similarly, YOLOv7-L+CBAM consistently maintains high precision scores, ranging from 0.74 to 0.90, across distance tests. The three highly performed detectors showcase reliability in identifying damages from various distances to the target area but trail the performance of Guided-DetNet by marginal differences. EfficientDet-D5+PAM, YOLOX-L+CBAM, and YOLOv5-m+CBAM continued to be the detectors that performed moderately compared to Guided-DetNet. EfficientDet-D5+PAM attained precisions ranging from 0.69 to 0.84, and YOLOX-L+CBAM recorded precisions ranging from 0.69 to 0.82. YOLOv5-m+CBAM precision ranges from 0.66 to 0.78 across the three varying distances. While the moderately performed detectors did not outperform the top detectors, their consistent accuracy suggests suitability for scenarios with varying distances during drone inspections.
Lastly, the least performed detectors, CSPFaster-R-CNN+EEAM+ and ResNet-34-VFNet+EEAM, attained precisions ranging from 0.62 to 0.74 and 0.59 to 0.71, respectively, across the three distances. The least performed detectors face challenges in maintaining high accuracy, particularly at longer distances, indicating limitations in capturing fine details from longer distances.

Overall, the distance-to-detection analysis shows the importance of selecting a detector that can maintain accuracy across varying distances during drone inspections. Guided-DetNet emerges as the top-performing detector, closely followed by YOLOv8-m-det+PAM, EnsembleDetNet, and YOLOv7-L+CBAM. Guided-DetNet and the highly performed detectors demonstrate the potential for effective structural damage detection across short, medium, and long distances, establishing a strong foundation for their real-world application in diverse scenarios.

*5.3.2 Detection under varying drone speed analysis*

The analysis of the detectors' performance under different speed scenarios provides valuable insights into their ability to detect structural damages accurately during dynamic flight conditions. The precision scores obtained under normal (10 km/h), moderate (20 km/h), and high (30 km/h) speeds are crucial for evaluating the robustness of each model.

Guided-DetNet emerges as the top-performing model across all speed tests, as seen in Fig. 16b. Even under high-speed conditions, Guided-DetNet maintains a commendable precision score of 0.81. This suggests that Guided-DetNet excels in detecting structural damages swiftly, as evidenced by its precision score range, which is from 0.81 to 0.89. This showcases Guided-DetNet's robustness in real-world applications where dynamic flight patterns are common. YOLOv8-m-det+PAM, EnsembleDetNet, and YOLOv7- L+CBAM exhibit competitive performance, closely following Guided-DetNet in precision scores across normal, moderate, and high-speed tests. YOLOv8-m-det+PAM maintains a consistently high level of precision, making it a reliable detector at varying speeds, with precision scores ranging from 0.78 to 0.87. EnsembleDetNet and YOLOv7-L+CBAM demonstrates reliability in detecting damages with precision scores ranging from 0.75 to 0.84 but with variation under moderate speed. In terms of moderately performed detectors, EfficientDet-D5+PAM, YOLOX-L+CBAM, and YOLOv5-m+CBAM demonstrate competitive precision across speed variations; however, they are slightly lower compared to the top-performing detectors. The performance decline under high-speed conditions is relatively moderate, indicating the moderately performed detectors' stability in dynamic environments to an extent. EfficientDet-D5+PAM attained a precision score ranging from 0.68 to 0.74, and YOLOv5-m+CBAM recorded precisions between 0.66 and 0.73, inclusive. Among the moderately performed detectors, YOLOX-L+CBAM attained better precision scores ranging from 0.70 to 0.82, as seen in Fig. 16b. CSPFaster-R-CNN+EEAM+ and ResNet-34-VFNet+EEAM consistently records the lowest precision scores across speed tests, ranging from 0.60 to 0.69, and 0.58 to 0.67, respectively. These detectors exhibit challenges in maintaining precision, particularly under high-speed conditions, highlighting limitations in handling dynamic flight patterns.

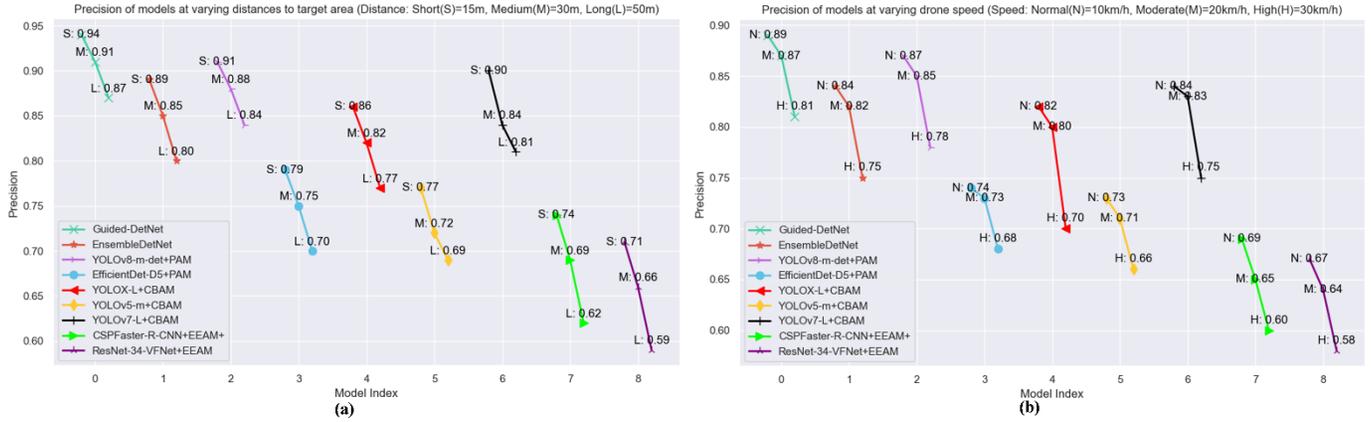

**Fig. 16.** Precision of Guided-DetNet and the compared detectors in the two setups of the filed test: (a) detection precision at varying distances to target area, and (b) detection precision at varying drone speed.

The speed test analysis emphasizes the significance of detectors that can maintain precision under dynamic flight conditions with varying speeds. Guided-DetNet stands out as the most robust detector, showcasing the potential for effective structural damage detection during drone flights at varying speeds.

Overall, regardless of the promising performance of the deployed drone with Guided-DetNet in the real-world experiment, adverse weather conditions, such as strong wind or rain, may impact drone stability; hence, integrating advanced stabilization mechanisms in drones can enhance their reliability in damage detection. Also, the applicability of Guided-DetNet with drones can be extended to industrial facilities like chimneys, silos, and power plants for detecting surface degradation, such as corrosion or cracks, during regular inspections. However, high temperatures, electromagnetic interference, and inaccessible areas can complicate inspections. As a result, drones with thermal shielding and advanced communication protocols, combined with Guided-DetNet's robust detection algorithms, can enable reliable inspections. Lastly, tunnels require regular inspections for surface deformations, cracks, and water seepage. Guided-DetNet's generative attention module (GAM) can be particularly effective in distinguishing foreground-background features in dimly lit environments, enabling precise damage localization. However, confined spaces may reduce the effectiveness of aerial drone navigation. Hence, deploying ground-based robots equipped with cameras and lights, paired with Guided-DetNet's detection capabilities, can overcome these challenges.

### 5.4 Ablation study

The ablation study conducted on Guided-DetNet involves the removal of two key modules, the Generative Attention Module (GAM) and the Hierarchical Elimination Algorithm (HEA), resulting in two ablated models: (1) Guided-DetNet-GAM, and (2) Guided-DetNet-HEA. This study aims to evaluate the impact of these modules on the overall performance of Guided-DetNet in structural damage detection.

From Fig. 17a, Guided-DetNet demonstrates superior precision (0.94), recall (0.86), and F1 score (0.90), indicating its effectiveness in accurately identifying and classifying damaged structural components. The removal of GAM in Guided-DetNet-GAM leads to a reduction in precision, recall, and F1 score, suggesting that GAM contributes significantly to the framework's ability to detect damaged areas correctly. Guided-DetNet-HEA, without HEA, maintains high precision, recall, and F1 score but trails that of Guided-DetNet's results. The results attained by Guided-DetNet-HEA indicate that HEA's contribution to the precision, recall, and F1 score performance of Guided-DetNet is minimal compared to the contribution of GAM. Regarding mAP, mAR, and AP, the removal of GAM in Guided-DetNet-GAM results in a slight decrease in AP, mAP, and mAR, implying that GAM contributes to improving the precision and accuracy of localization. Guided-DetNet-HEA shows a similar trend, with a decrease in AP, mAP, and mAR compared to the original Guided-DetNet, as seen in Fig. 17b. This suggests that both modules, GAM and HEA contribute to Guided-DetNet with varying improvements to the overall performance. Lastly, regarding FPS and inference time, Guided-DetNet achieves a commendable FPS of 57.04 and an inference of 6.22ms. However, the removal of GAM and HEA in Guided-DetNet-GAM and Guided-DetNet-HEA results in improved processing speed, with FPS increasing from 61.10 to 67.34, respectively. The inference time improves as well, with Guided-DetNet-HEA exhibiting the fastest inference time at 4.49 ms, as seen in Fig 17c. The improvement in inference time suggests that the removed modules contribute to computational overhead. The results indicate that both GAM and HEA contribute significantly to the detection performance of Guided-DetNet. While their removal leads to a decrease in certain metrics, it results in notable improvements in processing speed and reduced inference time. Table 4 shows the summary of the results of the ablation study, which complements Fig. 17.

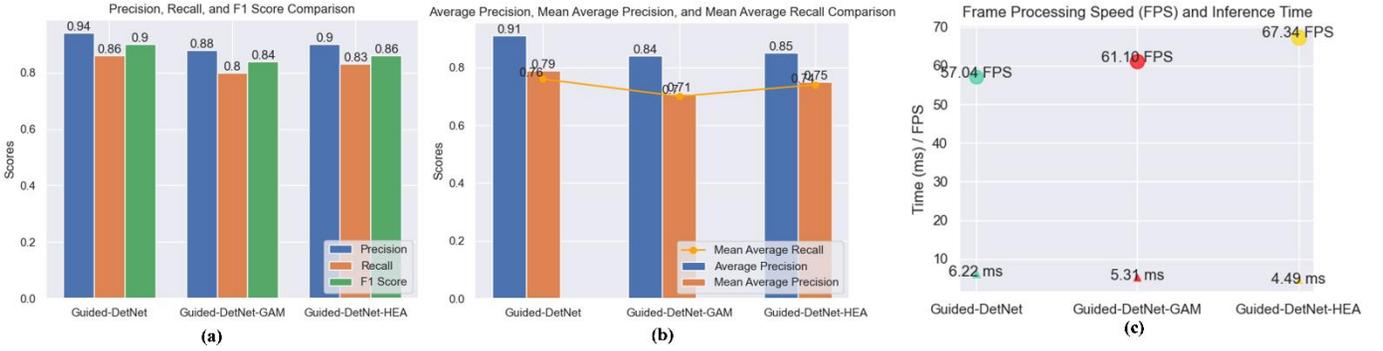

**Fig. 17.** Ablation study results for Guided-DeNet and the two ablated versions of Guided-DetNet.

Table 4
Numerical summary of ablation results for Guided-DetNet, Guided-DetNet-GAM, and Guided-DetNet-HEA

| Model | Precision | Recall | F1 Score | AP | mAP | mAR | FPS | Inference time |
|---|---|---|---|---|---|---|---|---|
| Guided-DetNet | 0.94 | 0.86 | 0.90 | 0.91 | 0.79 | 0.76 | 57.04 | 6.22 |
| Guided-DetNet-GAM | 0.88 | 0.80 | 0.84 | 0.84 | 0.71 | 0.70 | 61.10 | 5.31 |
| Guided-DetNet-HEA | 0.90 | 0.83 | 0.86 | 0.85 | 0.75 | 0.74 | 67.34 | 4.49 |

In conclusion, the ablation study provides valuable insights into the importance of GAM and HEA in Guided-DetNet. The analysis allows for informed decisions regarding framework configuration based on the priorities of speed and accuracy in structural damage detection applications.

## 6. Conclusions

In this study, Guided-DetNet, which comprises a classifier and a detector, is presented for triple classification and dual detection structural health monitoring tasks. The main contributions of Guided-DetNet include three main key modules: (1) Generative Attention Module (GAM), (2) Hierarchical Elimination Algorithm (HEA), and (3) Volumetric Contour Visual Assessment (VCVA). GAM utilizes cross-horizontal and cross-vertical patch merging and cross foreground-background feature fusion to generate multi-varied feature maps to increase the diversity of feature maps, thereby increasing the robustness of the framework. GAM enhances the framework's ability to focus on relevant regions in an image, contributing to accurate structural damage detection. The HEA is a class instance refinement algorithm that utilizes hierarchical relationships among various class instances in the inter-task to mitigate noisy labels associated with data labels by eliminating unlikely class instances given an image. Lastly, the VCVA is a quantification module that aids in understanding the severity of detected damages. The VCVA provides a detailed 3D volumetric representation of the detected damage with associated intensity, as well as a score, which determines the surety of the 3D volumetric representation of the detected damage.

To better gain insight into the performance of Guided-DetNet, a triple classification and dual detection experiment was conducted, where Guided-DetNet was compared with several current state-of-the-art models. In the triple classification task, which comprised scene level, collapse mode, and component type classification, Guided-DetNet recorded an accuracy of 96%, a precision of 1.0, a recall of 0.92, and a misclassification rate of 0.04. Comparing Guided-DetNet to the best-compared classifier, YOLOv8-m-cls+PAM, in the triple classification task, Guided-DetNet was better by a margin of 2% to 3% in most metrics. In the dual detection task, which comprised damage level and damage type, Guided-DetNet performed better than the compared detectors. In a general quantitative analysis, Guided-DetNet recorded a marginal difference of approximately 3% higher than the best-compared detector, YOLOv8-m-det+PAM, in metrics such as precision, recall, mAP, and mAR. Other competitive detectors, including YOLOv7-L+CBAM and EnsembleDetNet, attained promising performance across varying evaluation metrics but lagged behind Guided-DetNet. A comparison of fps and inference time revealed that Guided-DetNet was competitive, with an fps of 57.04 and an inference time of 6.22m/s, which makes it suitable for real-time applications. Although Guided-DetNet was not the best in terms of fps and inference time, Guided-DetNet offered a balance between accuracy and efficiency.

Under dual detection, further experiments categorized under environmental robustness tests and noisy labeled tests were carried out to test the resilience of Guided-DetNet and the compared detectors. In the environmental robustness tests, Guided-DetNet was tested under (1) images with background variations, (2) occluded images, and (3) simulated rain-snowy images. Across all three environmental robustness tests, Guided-DetNet remained resilient by reaching precisions ranging from 0.79 to 0.92 under varying intensities. The best-compared detector, YOLOv8-m-det+PAM, recorded precisions ranging from 0.77 to 0.90. In the second robustness test, training and testing with noisy labeled data, Guided-DetNet attained an accuracy of 95% while YOLOv8-m-det+PAM recorded 91%. Additionally, Guided-DetNet offered a 3D volumetric representation and quantification of the severity of detected damages to gain more insight regarding assessment. Lastly, two types of field tests comprising varying distances to the detection area and detection under varying drone speeds were carried out to assess Guided-DetNet in terms of drone-based structural health monitoring. Guided-DetNet outperformed the compared detectors in the field test by a marginal difference of at least 2%.

In conclusion, Guided-DetNet emerged as a robust and accurate framework for structural damage detection, offering a fine balance between accuracy and efficiency. The quantitative comparative analyses, robustness tests, and field tests collectively contribute to the understanding of Guided-DetNet's strengths and potential areas for optimization, which was revealed in an ablation study. This study potentially lays a foundation for advancing the field of structural damage detection and holds promise for practical applications, particularly in the domain of drone-based structural health inspections. However, modules like HEA in Guided-DetNet can be optimized for faster inferences. In the near future, a possible direction to look into is the integration of advanced sensors to explore hyperspectral or thermal features, which can reveal hidden structural defects to improve performance under diverse environmental conditions. Also, incorporating dynamic structural behaviors such as vibration and load responses into the detection of structural damages for a comprehensive SHM will be explored. Additionally, the adaptability of Guided-DetNet offers potential applications in other domains. For instance, Guided-DetNet's robust feature extraction and hierarchical classification capabilities could be adapted for medical imaging tasks, where precise localization and quantification of anomalies are critical. Also, the framework's ability to handle noisy data and complex environments could be leveraged for detecting land use changes, crop health, or natural disasters using satellite or aerial imagery in remote sensing tasks.

**Acknowledgments**

The authors thank Priscilla Fosuah Sarkodie, Ruth Dwomoh Agyemang, and Abigail Boamah for extensive proofreading. Last but not least, this study is dedicated to the memory of Dr. Aaron Opoku Agyemang, who passed away on the 27$^{th}$ of July, 2022, forever in our hearts.